\documentclass[lettersize,journal]{IEEEtran}
\usepackage{amsmath,amsfonts,amssymb}
\usepackage{algorithmic}
\usepackage{algorithm}
\usepackage{array}
\usepackage{textcomp}
\usepackage{stfloats}
\usepackage{url}
\usepackage{verbatim}
\usepackage{graphicx}
\usepackage{cite}
\usepackage{multicol}

\usepackage[utf8]{inputenc}
\usepackage[T1]{fontenc}

\usepackage{mf-style}
\usepackage{agent-symbols}

\usepackage{color}

\renewcommand{\mfr}[1]{\textcolor{red}{#1}}
\renewcommand{\mfb}[1]{\textcolor{black}{#1}} % for publication
\usepackage{multirow}
\usepackage[]{subcaption}
\usepackage{blkarray} % matrix with labels

\usepackage{listings}
\begin{document}

\title{
Simplification of Robotic System Model Analysis by Petri Net Meta-Model Property Transfer
}

\author{Maksym Figat$^{*}$,~\IEEEmembership{Member,~IEEE} and Cezary Zieli{\'{n}}ski, ~\IEEEmembership{Senior Member,~IEEE}% <-this % stops a
	% <-this % stops a space
	\thanks{Maksym Figat and Cezary Zieli{\'{n}}ski are with Warsaw University of Technology, Institute of Control and Computation Engineering, Warsaw, Poland. $^{*}$Corresponding author: Maksym Figat (\tt maksym.figat@pw.edu.pl)}
	\thanks{Research was funded by the Warsaw University of Technology within the Excellence Initiative: Research University (IDUB) programme.}
	\thanks{This work has been submitted to the IEEE for possible publication. Copyright may be transferred without notice, after which this version may no longer be accessible.}
}

\maketitle

\begin{abstract}
This paper presents a simplification of robotic  system model analysis due to the transfer of Robotic System Hierarchical Petri Net (RSHPN) meta-model properties onto the model of a designed system. Key contributions include: 1) analysis of RSHPN meta-model properties; 2) decomposition of RSHPN analysis into analysis of individual Petri nets, thus the reduction of state space explosion; and 3) transfer of RSHPN meta-model properties onto the produced models, hence elimination of the need for full re-analysis of the RSHPN model when creating new robotic systems. Only task-dependent parts of the model need to be analysed. This approach streamlines the analysis thus reducing the design time. Moreover, it produces a specification which is a solid foundation for the implementation of the system.
The obtained results highlight the potential of Petri nets as a valuable formal framework for analysing robotic system properties.

\end{abstract} 

\begin{IEEEkeywords}

Robotic system design, Hierarchical Petri Net, Robotic system meta-model, Model analysis
\end{IEEEkeywords}

% !TeX root = figat_analiza.tex

\section{Introduction}
\label{sec:introduction}

As robots increasingly execute complex tasks in close proximity to humans, ensuring their accurate and safe operation is crucial. However, formal methods for system description, a vital component of model-based approaches, are often overlooked, leading to the detection of flaws as late as the implementation or even verification stages, thus resulting in costly redesigns~\cite{Brugali:2020}. Component-based development frameworks, e.g., ROS, offer flexibility but lack stringent guidelines, making the quality of the resulting systems heavily dependent on the designer's experience, often leading to inferior software code~\cite{Silva:2021}. The quality of such systems can be verified at the runtime verification stage, i.e., there are many ROS-based tools for analysis, e.g., SOTER~\cite{Shivakumar:2020}, ROSMonitoring~\cite{Ferrando:2020}, and ROSRV~\cite{Huang:2014}. However, these tools primarily analyse message flow between nodes treated as black boxes, making it impossible to assess the state of subsystems directly. This significantly complicates the overall analysis of the designed system and narrows the space of properties that can be examined, as stated in~\cite{DalZilio:2023}. \mfb{The utilization of formal methods could greatly improve system reliability and safety, minimizing risks and redesign expenses. Among these, Petri nets (PN)~\cite{Murata:1989:Petri:Nets} have been widely adopted for modelling and verifying concurrent and distributed systems due to their expressiveness in representing synchronization, resource allocation, and event-driven behaviour.}

Formal methods are necessary to perform model (i.e., specification) checking before implementation. They play a crucial role in model-based design methods, enabling system specification~\cite{sparc:2016} and analysis, though their application to robotics remains a challenge~\cite{Silva:2021}. Many model-based approaches focus on specific robot tasks (e.g., multi-robot path planning~\cite{Hustiu:2024,He:2024}) or particular system layers (e.g., decision layer~\cite{Lesire:2020:robot-skills} and functional layer~\cite{DalZilio:2023}), without providing a holistic view of the system. Moreover, many model-driven engineering approaches emphasize implementation while sidelining specification. This results in a tight coupling between modelling and implementation concepts, hindering the development of platform-independent models and comprehensive system analysis~\cite{Silva:2021}.

\mfb{One of the problems with formal system models is their complexity resulting in: (a) the high computational cost of formal analysis~\cite{Su:2023}, (b) scalability problems in model checking~\cite{DalZilio:2023}, and (c) explosion of the state space, particularly in systems with variable structures, such as multi-robot systems modelled by Variable Petri Nets (VPNs)~\cite{Su:2023}.}

\mfb{Several methods have been proposed to address these challenges. One approach focuses on optimizing transition grouping within PNs~\cite{Dou:2024}, where Maximally Good-Step (MG) graphs are introduced to reduce state-space explosion compared to traditional reachability graphs. While this method improves transition management, it still depends on explicit state-space exploration. In~\cite{Yang:2022} VPNs are transformed  into Kripke structures, enabling formal property verification through model checking. However, despite its rigour, this approach remains constrained by the inherent state-space size issues.}

\mfb{For multi-agent systems,\cite{He:2023} employs Knowledge-Oriented Petri Nets (KPNs) with Reduced Ordered Binary Decision Diagrams (ROBDD) for privacy-critical verification, significantly compressing state representations. Nonetheless, this approach still relies on full state enumeration. Similarly,\cite{Dong:2024} introduces a symbolic observer-based method for verifying state opacity in networked discrete event systems, mitigating some aspects of state-space explosion but ultimately requiring extensive reachability analysis. An alternative strategy for managing large PNs is the use of reduction equations, as demonstrated in~\cite{Berthomieu2020}, where a compact representation of the state space is achieved. While effective for counting markings, this method remains dependent on reachability analysis.}

\mfb{A possible way to overcome these limitations is through hierarchical modelling, which reduces complexity by structuring the system into levels of abstraction. Such methods have been proposed in the context of PN~\cite{Huber91,Valk:2003,Vogler:1992,Zuberek:1996,Luo:2015}, allowing local analysis of subsystems while maintaining global system properties. However, many existing hierarchical PN (HPN) models introduce additional structural complexity due to unrestricted substitution mechanisms, what complicates property verification~\cite{Murata:1989:Petri:Nets,Luo:2015,Kim:2018}. Moreover, these models often allow transitions and places with multiple input and output connections, making it difficult to apply reduction techniques while preserving system properties. Some hierarchical approaches introduce place fusion mechanisms~\cite{Szpyrka:08}, yet they lack constraints ensuring modular analysis. As a result, while HPNs improve scalability compared to flat PNs, their structural complexity still poses challenges for formal analysis. Consequently, additional constraints and decomposition mechanisms are needed to enhance HPN for robotic system modelling and verification.}

%%%%%%%%%%%%%%%%%%%%%%%%%%%%%

\mfb{Decomposition of the analysis is often seen as a remedy to the problem of excessive model complexity~\cite{Ding:2019}. However, many decomposition approaches are limited to basic applications, such as motion planning in simple robots operating in fixed environments. Another widely used technique to mitigate state-space explosion is runtime verification (RV), which does not require a formal model of the system but significantly restricts the set of properties that can be analysed. A refinement of this approach, known as predictive RV~\cite{Pinisetty:2017,Pelletier:2023}, incorporates a model of the system as \emph{a priori} knowledge, enabling future state prediction. However, this still occurs only after the implementation phase, not in early design verification.}

\mfb{To avoid these limitations, a HPN model tailored to robotic systems was proposed in \cite{Figat:2022:RAS}, i.e.\ Robotic System Hierarchical Petri Net (RSHPN). The parametric RSHPN meta-model~\cite{Figat:2022:RAS} is a six-layered HPN modelling the activity of the entire robotic system. Parameters derived from robotics ontology define both the structure and the activities of the system and its components.
RSHPN introduces structural constraints that enable systematic decomposition while maintaining analytical tractability. Unlike previous hierarchical approaches, RSHPN uses only single-input/single-output place nets (i.e.\ workflow nets), enabling modular analysis. While \cite{Figat:2022:RAS} introduced the RSHPN meta-model, it did not investigate its formal properties or its advantages in system analysis. This article fills this gap by presenting a detailed study of RSHPN’s properties and demonstrating its benefits for robotic system design. It shows how RSHPN supports property transfer, structural decomposition, and invariant-based analysis, making it a practical tool for modeling and verifying robotic systems.} Moreover, a domain-specific Robotic System Specification Language (RSSL)\cite{Figat:2022:RAL} can be used to define RSHPN parameters, transforming the meta-model into an executable model. This model is then used to generate controller code and perform system analysis. This process forms the basis of the Robotic System Specification Methodology (RSSM)\cite{Figat:2020:Access}\footnote{Introduction to the RSSM: https://youtu.be/027JvJ-CtjY}.
Conventional PN analysis tools (e.g., Tina~\cite{Tina:Tool:LAAS:www}) applied to a flattened-out RSHPN were of limited utility, even for a simple robotic system~\cite{Figat:2022:RAS}, due to state-space explosion. A PN with 1168 places, 1294 transitions, and 3085 edges produced a reachability graph exceeding 25 million nodes and 163 million edges. It was not possible to analyse this model using a machine with 32 GB RAM. For more complex systems, such as multi-robot systems with dynamic structures~\cite{Yang:2023}, this problem would only be aggravated. Thus, an alternative method for analysing RSHPN was devised. This paper demonstrates that the RSHPN structure significantly simplifies system analysis through hierarchical decomposition (video tutorial\footnote{RSHPN analysis video: https://youtu.be/JenAB1IKVVY}).

Paper organisation is as follows:
\mfb{Sec.~\ref{sec:contribution}~--~contribution of this article,}
Sec.~\ref{sec:eaa}~--~introduction to the robotic system architecture based on embodied agents,
Sec.~\ref{sec:introduction_to_petri_nets}~--~introduction to PNs,
\mfb{Sec.~\ref{sec:pn-properties-to-be-analysed}~--~properties to be verified,
Sec.~\ref{sec:pn-analysis-methods}~--~the analysis methods,}
Sec.~\ref{sec:rshpn_metamodel}~--~RSHPN meta-model,
Sec~\ref{sec:rshpn_model_analysis}~--~RSHPN meta-model analysis,
Sec.~\ref{sec:results}~--~analysis results,
Sec.~\ref{sec:case-study}~--~case study based on exemplary robotic system,
Sec.~\ref{sec:discussion}~--~discussion,
and Sec.~\ref{sec:conclusions}~--~conclusions. % OK

% !TeX root = ../figat_analiza.tex

\section{\mfb{Contribution}}
\label{sec:contribution}
%%%%%%%%%%%%%%%%%%%%%%%%%%%%%%%%%%%%%%%%%%%%%%%%

\mfb{The paper focuses on formal analysis of  RSHPN metamodel  based robotic systems~\cite{Figat:2022:RAS}. The main contributions are:\\
\textbf{1) Analysis Techniques for RSHPN:} Unlike prior work, we develop a systematic methodology for verifying RSHPN properties of the meta-model, ensuring safety, conservativeness, and liveness without requiring exhaustive reachability analysis. This formal verification reduces computational overhead compared to classical PN-based methods~\cite{Murata:1989:Petri:Nets, Luo:2015, Kim:2018}.\\
\textbf{2) Hierarchical Decomposition for Scalable Analysis:} We introduce a RSHPN decomposition method that enables independent analysis of subnets while maintaining global system correctness. This significantly mitigates state-space explosion.\\
\textbf{3) Meta-Model Property Preservation in the Generated Models:} We show that properties verified for the RSHPN meta-model are automatically transferred to specific models. As a result, full system re-analysis is unnecessary; only the task-dependent and inter-subsystem communication fragments require verification, what improves scalability.\\
\textbf{4) Advancement Beyond Prior Work:} In contrast to~\cite{Figat:2022:RAS}, which defined the RSHPN meta-model, this paper provides a complete analysis methodology. Additionally, while~\cite{Figat:2022:RAL} introduced the RSSL for defining RSHPN parameters, it did not propose analysis techniques. Here  formal verification methods applicable to RSSL-generated RSHPN models are introduced.}
\mfb{These contributions establish RSHPN as a scalable and structured framework for modelling and verifying robotic systems, addressing key challenges in formal system analysis.}

  % OK

 %!TeX root = ../figat_analiza.tex

\section{Robotic System Architecture}
\label{sec:eaa}

\begin{table}[tb]
	\centering
	\footnotesize
	%	\scriptsize
	\setlength{\belowdisplayskip}{0pt} \setlength{\belowdisplayshortskip}{0pt}
	\setlength{\abovedisplayskip}{0pt} \setlength{\abovedisplayshortskip}{0pt}	
	\setlength{\tabcolsep}{2pt} % space between columns
	
	\caption{Robotic System Architecture notation}
	\scalebox{0.97}{ % {0.75}{	
		%%%%%%%%%%%%%%%%%%%%%%%%%%%%%%%%%%%%%%%%%%%%%%%%%%%%	
		\begin{tabular}{c|l|c|l}
			%			\hline
			\multicolumn{4}{c}{\includegraphics[width=1.0\linewidth]{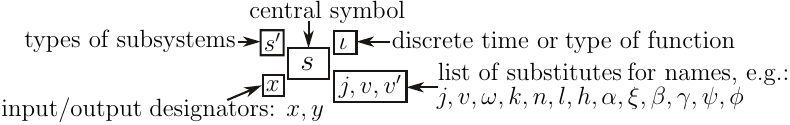}}\\
			%%%%%%%%%%%%%%%%%%%%%%%%%%%%%%%%%%%%%%%%%%%%%%%%%%%%
			\hline
			\multicolumn{2}{c|}{\bf Structure:} & \multicolumn{2}{c}{\bf Activity:}\\ \hline
			Central & Description & Central  & Description\\
            symbol  &             & symbol   &      \\ \hline
			$\agentX{}$ & agent & 	\subsystemTaskX{} & task \\
			$C$ & control system & \behaviourXXX{}{}{} & behaviour \\
			\realEffectorX{} & real effector & \transitionFunctionXX{}{} & transition function\\
			\realReceptorX{} & real receptor & \initialConditionXX{}{} & initial condition\\
			\subsystemX{} & subsystem & \terminalConditionXX{}{}& terminal condition \\
			\controlSubsystemX{} & control subsystem & \errorConditionXX{}{} & error condition\\
			\virtualEffectorX{} & virtual effector &\interAgentCommunicationChannelX{} & inter-agent channel\\
			\virtualReceptorX{} & virtual receptor & \intraAgentCommunicationChannelX{} & intra-agent channel\\
			& & & \\
            \hline
			%%%%%		
			List of & Description & List of & Description\\
            names   &             & names   & \\
            \hline
			\agentName, \agentName' & names of agents & \initialConditionName & name of initial condition\\
			\subsystemName, \subsystemName' & names of subsystems & \terminalConditionName & name of terminal condition\\	
			\virtualReceptorName & name of virtual receptor & \errorConditionName & name of error condition\\
			\virtualEffectorName & name of virtual effector & \transitionFunctionName & name of transition function\\
			\realReceptorName & name of real receptor & \partialFunctionName & name of partial function\\
			\realEffectorName & name of real effector & \overloadedFunctionName & name of overloaded function\\
			$v$& subsystem name & \behaviourName, \behaviourName' & names of behaviours\\
			
			%%%%%%%%%%%%%%%%%%%%%%%%%%%%%%%%%%%%%%%%%%%%%%%%%%%%
		\end{tabular}
	}
	%%%%%%%%%%%%%%%%%%%%%%%%%%%%%%%%%%%%%%%%%%%%%%%%%%%%	
	\label{table:agent_name_symbols_structure_activity}
\end{table}

\begin{figure}
	\begin{subfigure}{0.58\linewidth}
		\centering
		\includegraphics[width=1.0\linewidth]{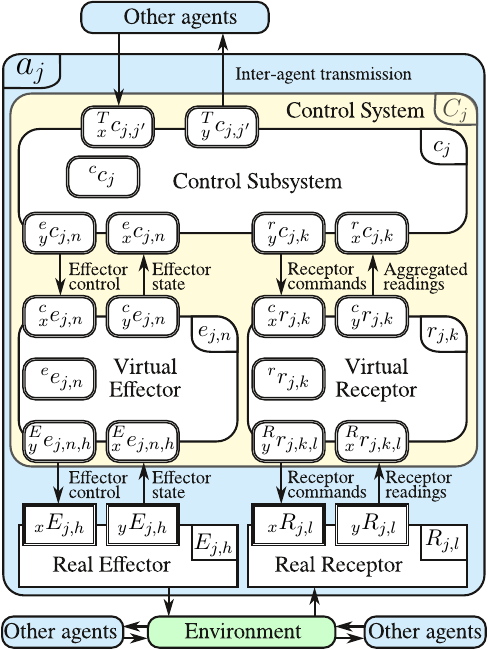}
		\caption{}
		\label{fig:agentstructureextended}
	\end{subfigure}
	\begin{subfigure}{0.41\linewidth}
		\centering
		\includegraphics[width=1.0\linewidth]{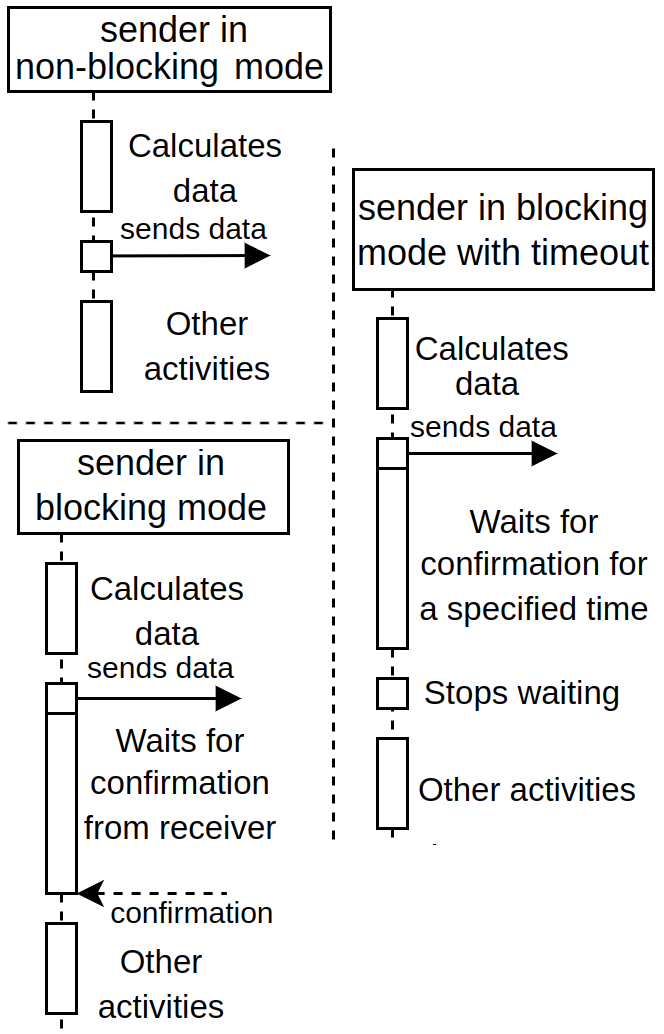}
		\caption{}
		\label{fig:sender_modes}
	\end{subfigure}
	\caption{(a): Structure of an embodied agent; (b): 3 communication modes from the perspective of the sending subsystem}
	\label{fig:embodied_agent_and_communication}
\end{figure}

{\bf Structure of a~robotic system:} A robotic system \roboticSystem\ consists of one or more robots and auxiliary devices. Each robot consists of one or more embodied agents~\cite{Figat:2022:RAS}. An embodied agent \agentj{j} (Fig. ~\ref{fig:agentstructureextended}) consists of: a single control subsystem \cbbbj{j}, real receptors \Rbbbj{j,l}, real effectors \Ebbbj{j,h}, virtual effectors \ebbbj{j,n}, and virtual receptors \rbbbj{j,k}. Tab.~\ref{table:agent_name_symbols_structure_activity} provides the description of the notation used. Receptors $\Rbbbj{j,l}$ collect data from the environment and pass it to $\rbbbj{j,k}$ to be aggregated and subsequently sent to $\cbbbj{j}$. Based on this data and the task, $\cbbbj{j}$ produces control commands for $\ebbbj{j,n}$, which in turn transforms them to affect the environment via $\Ebbbj{j,h}$.
Subsystems share a common structure. Each subsystem $\subsystemX{j,v}$ of an agent $\agentX{j}$ contains:
1)~$\internalMemoryX{j,v}$ (internal memory), 2)~input buffers $\inputBufferSetXXX{j,v}{}{}$, and 3)~output buffers $\outputBufferSetXXX{j,v}{}{}$ (where $\hat{X}$ is a set of X).

{\bf Activity of a~robotic system:} A robotic system task \subsystemTask\ is composed of the tasks \subsystemTaskX{j} executed by the agents \agentj{j}, and those are composed of tasks \subsystemTaskX{j,v}
executed by subsystems \subsystemX{j,v}. Subsystem tasks \subsystemTaskX{j,v} are executed by
selecting specific behaviours \behaviourX{j, v, \omega}.
The specific behaviour to be executed is selected by testing its initial condition \initialConditionX{j,v,\alpha} ($\alpha$ --
predicate designator)~\cite{Figat:2022:RAS}.
Each behaviour iteratively:
%\begin{enumerate}
1)~Calculates the associated transition function \transitionFunctionX{j,v,\transitionFunctionName}, which takes as arguments the data from internal memory \internalMemoryXX{j,v}{\iota} and input buffers \inputBufferXX{j,v}{\iota}, and inserts the calculated results into the output buffers \outputBufferXX{j,v}{\iota+1} and the internal memory \internalMemoryXX{j,v}{\iota+1}, where $\iota$ is the current time instant and $\iota+1$ is the next time instant: $\left( \internalMemoryXX{j,v}{\iota+1}, \, \outputBufferXX{j,v}{\iota+1} \right) :=\ \transitionFunctionX{j,v,\transitionFunctionName} \left(\internalMemoryXX{j,v}{\iota},\, \inputBufferXX{j,v}{\iota}\right)$,
2)~Sends data from the output buffers \outputBufferXX{j,v}{\iota+1} to the associated subsystems,
3)~Increments the discrete time counter $\iota$,
4)~Receives data from the associated subsystems and inserts it into input buffers $\inputBufferXX{j,v}{\iota}$, and
5)~Checks the terminal \terminalConditionX{j,v,\xi} and error \errorConditionX{j,v,\beta} conditions.
%\end{enumerate}
Behaviour \behaviourXXX{j,v,\omega}{s}{} terminates when one of those conditions is fulfilled.
In such a~case the \sbbbj{j,v} selects the next behaviour, e.g. \behaviourXXX{j,v,\omega'}{s}{} based on the fulfilled initial condition \initialConditionXX{j,v,(\omega,\omega')}{s} defined by the subsystem task \subsystemTaskX{j,v}.

Two subsystems, $s_{j,v}$ and $s_{j',h}$, can communicate, each using one of three modes: 1) blocking, 2) blocking with timeout, or 3) non-blocking mode. \mfb{As those modes pertain to sending and receiving, 9 communication models result~\cite{Figat:2019:ICRA,Figat:2022:RAS}.} In blocking mode $s_{j,v}$ waits for $s_{j',h}$ to acknowledge receiving data, while in non-blocking mode, $s_{j,v}$ continues its activities after sending data, as presented in Fig.~\ref{fig:sender_modes}. In blocking mode $s_{j',h}$ waits for data, while in non-blocking mode it continues its tasks even if data is not available.

% ====  OK ======
% !TeX root = ../figat_analiza.tex

%\clearpage

\section{Petri net preliminaries}
\label{sec:introduction_to_petri_nets}

\subsection{Petri Net (PN)}
PN is a bipartite graph with two kinds of nodes: places $\pnPlace{}{}$ (single circles) and transitions \pnTransitionXXX{}{}{} (rectangles). Directed arcs connect places to transitions and vice versa. PNs are represented either graphically (e.g.\ Fig.~\ref{fig:hpnnowauproszczona}) or algebraically.
Evolution of PN marking represents the changes of system state.
$\mu(\pnPlaceX{i})$ indicates marking, i.e.\ number of tokens (filled-in black circles) in a place $\pnPlaceX{i}\in \hat{\pnPlaceX{}}$, where $\mu(p_i)\ge0$ and $\hat{\pnPlaceX{}}$ is a set of places;
$M_m$ is a PN marking vector in instant $m$, $M_m=\big(\mu^{m}(\pnPlaceX{1}),\ldots,\mu^{m}(\pnPlaceX{k})\big)^{\top}$, where $k=\setSize{\hat{p}}$ and $m\ge 0$ (in general $|\hat{\star}|$ is the cardinality of set $\hat{\star}$);
$M_{0}$ indicates the initial marking and $M_m$ marking $m$ derived from $M_{0}$;
$M_m(p_i)$ is the number of tokens in place $p_{i}$ while in $M_m$. %=\mu^{m}(p_i)$.
PN in an algebraic form is represented by the incidence matrix. For a~PN with $\setSize{\hat{t}}$ transitions and $\setSize{\hat{p}}$ places, the incidence matrix $N=[n_{ij}]$ is a $\setSize{\hat{t}} \times \setSize{\hat{p}}$ matrix containing integers; $n_{ij}$ represents the change of the number of tokens in a~place \pnPlaceX{j} when $\pnTransitionX{i} \in \hat{\pnTransitionX{}}$ fires~\cite{Murata:1989:Petri:Nets}:
\begin{equation}
	{\begin{array}{c}
			n_{ij} = n_{ij}^{+} - n_{ij}^{-}
		\end{array}
	}
	\label{eq:incidence_matrix_entry}
\end{equation}
$n_{ij}^{-}$ denotes the number of tokens removed from the input place \pnPlaceX{j} and $n_{ij}^{+}$ denotes the number of tokens inserted into  \pnPlaceX{j}, both after firing \pnTransitionX{i}.
If for $M_m$ $\forall \pnPlaceX{j} \; (n_{ij}^{-}\leq M_m(\pnPlaceX{j}))$, then \pnTransitionX{i} fires causing a transition: $M_m \xrightarrow{\pnTransitionX{i}} M_{m+1}$. The marking $M_m$ is a~column vector $\setSize{\hat{p}} \times 1$. As the $i$-th row in the incidence matrix $N$ denotes the change in the marking from $M_{m}$ to the marking $M_{m+1}$ as a~result of firing \pnTransitionX{i}, the new marking of the network can be expressed by the formula Eq.~(\ref{eq:firing_transition_equation}).
\begin{equation}
	{\begin{array}{c}
			M_{m+1} = M_{m} + N^{\top} \cdot u_{m+1}
		\end{array}
	}
	\label{eq:firing_transition_equation}
\end{equation}
where $u_{m+1}$ is a~column vector $\setSize{\hat{t}} \times 1$ containing  1 at position $i$ (denoting the firing of \pnTransitionX{i}) with 0 at all other positions. Thus:
\begin{equation}
	{\begin{array}{c}
			M_{m} - M_{0} = N^{\top} \cdot (u_{1} + \ldots + u_{m}) = N^{\top} \cdot \bf x
		\end{array}
	}	\label{eq:reachable_marking_from_initial_marking_5}
\end{equation}
where $\bf x$ (called firing count vector~\cite{Murata:1989:Petri:Nets}) is a~column vector of dimension $\setSize{\hat{t}} \times 1$ with non-negative integer values, where $i$-th value in the vector $\bf x$ denotes the number of times the transition \pnTransitionX{i} fires, leading from $M_0$ to $M_m$, i.e. $M_0 \xrightarrow{{\bf x}} M_{m}$.

%%%%%%%%%%%%%%%%%%%%%%%%%%%%%%%%%%%%%%%%%%%%%%%%%%%%%%%%%%%%%%%
\subsection{Hierarchical Petri Net (HPN)}
\label{subsec:hpn}
HPN \pnHierarchicalXXX{}{}{}~\cite{Figat:2019:ICRA,Figat:2022:RAS} extends a PN by introducing pages \pnPageXXX{}{}{} (double circles) representing an underlying PN (rectangular panel connected by a dashed arrow to its page -- Table~\ref{table:rshpn_notation_concept}). Each such net has a single input \pnInputPlace{} and a single output \pnOutputPlace{} place. When the token appears within page \pnPageXXX{}{}{}, it appears in the input place of the lower level PN (token ``drops'' down), and the lower PN is activated. When its activity is finished then the token appears within the output place \pnOutputPlace{} (token ``jumps up'' to the page), and becomes  visible to the PN containing the page. Logical conditions \pnCondition{} and priorities \pnPriority{} are associated with transitions, and operations $\pnOperation$ with places.

\begin{table}
	\caption{RSHPN notation~\cite{Figat:2022:RAS}}
	\centering
	\footnotesize
	%	\scriptsize
	\setlength{\belowdisplayskip}{0pt} \setlength{\belowdisplayshortskip}{0pt}
	\setlength{\abovedisplayskip}{0pt} \setlength{\abovedisplayshortskip}{0pt}	
	\setlength{\tabcolsep}{2pt} % space between columns
	
	\scalebox{0.8}{	
		%%%%%%%%%%%%%%%%%%%%%%%%%%%%%%%%%%%%%%%%%%%%%%%%%%%%		
		\begin{tabular}{|c|l||c|l||c|l|}
			\hline
			\multicolumn{4}{|c||}{Notation of RSHPN concepts} & \multicolumn{2}{c|}{Graphical notation}\\ \hline
			\multicolumn{4}{|c||}{\includegraphics[width=0.7\linewidth]{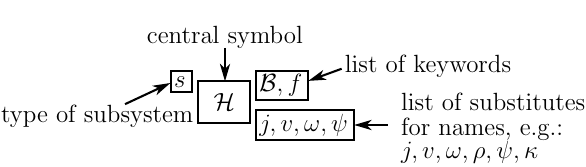}} &\multicolumn{2}{c|}{\includegraphics[width=0.45\linewidth]{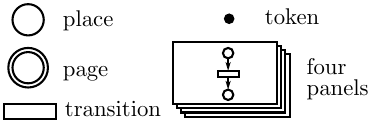}}\\ \hline \hline
			%%%%%%%%%%%%%%%%%%%%%%%%%%%%%%%%%%%%%%%%%%%%%%%%%%%%
			%%%%%%%%%%%%%%%%%%%%%%%%%%%%%%%%%%%%%%%%%%%%%%%%%%%%
			Central & Description & Substi- & Names: & Keyword & Description\\
			symbol & & tutes & & & \\ \hline
			\pnPlaceX{} & place & 	\agentName & agent & $\rm in$ & input place\\
			\pnPageX{} & page & \subsystemName & subsystem & $\rm out$ & output place\\
			\pnTransitionX{} & transition & \behaviourName & behaviour & \behaviourXXX{}{}{} & behaviour layer\\
			\pnHierarchicalX{} & hierarchical & $\psi$ & partial function & $\rm snd$ & send sublayer\\
			&  Petri net & $\rho$ & output buffer & $\rm rcv$ & receive sublayer\\
			\pnConditionX{} & condition  & $\kappa$ & input buffer & $\rm fusion$ & fused place\\
			\agentX{} & agent & & &$f$ & transition function\\
			\subsystemX{} & subsystem & &
			% \cline{1-4}
			% \multicolumn{4}{c|}{} &
			&  & sublayer\\
			\cline{1-6}
			%%%%%
			%			\hline
			%%%%%%%%%%%%%%%%%%%%%%%%%%%%%%%%%%%%%%%%%%%%%%%%%%%%
		\end{tabular}
	} % end scalebox
	%%%%%%%%%%%%%%%%%%%%%%%%%%%%%%%%%%%%%%%%%%%%%%%%%%%%	
	\label{table:rshpn_notation_concept}
\end{table}

\begin{figure}
	\centering
	\includegraphics[width=1.0\linewidth]{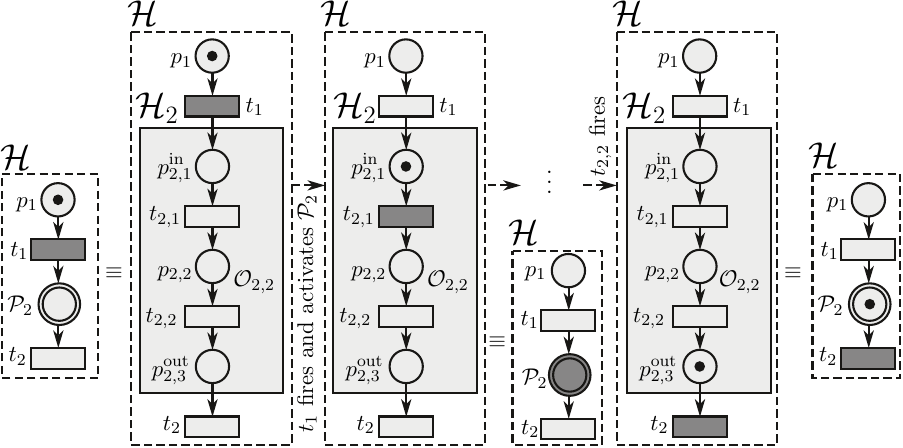}
	\caption{\mfb{Execution of a~HPN represented by a page (example)}}
	\label{fig:rshpn-petri-net-example}
\end{figure}

\mfb{\textbf{Firing of an enabled transition connected to a~page}. Darker shade represents active nodes (places, transitions, pages) in Fig.~\ref{fig:rshpn-petri-net-example}. Initially transition \pnTransitionX{1} is enabled. A place or a page can be in one of three states: 1)~not activated, 2)~activated (having darker colour) and 3)~completed (with a~single token). While the place or page is activated it executes either an operation associated with the place or activates of the HPN represented by the page. From the perspective of a~transition waiting for a~token from an activated place or a~page, the token will eventually appear and then the transition will become enabled.
 A token within the page \pnPageX{2} indicates that the represented HPN \pnHierarchicalX{2} has finished its activities, i.e.\ a~single token had appeared in its output place and so no transition is enabled within \pnHierarchicalX{2}. From the perspective of \pnHierarchicalX{} the page \pnPageX{2} can be viewed as a~place executing a~more complex operation represented by another HPN, i.e. \pnHierarchicalX{2}. The transition \pnTransitionX{2} will be eventually enabled, unless a~deadlock occurs within \pnHierarchicalX{2}.
}

\mfb{
\textbf{Fused places.}
\label{sec:rshpn-example-fusion-places}
If two places: $\pnPlaceX{\pnHierarchicalX{1},\alpha}$ and $\pnPlaceX{\pnHierarchicalX{2},\beta}$, belonging respectively to nets: \pnHierarchicalX{1} and \pnHierarchicalX{2}, are fused, they become a~single place in the combined net, as presented in Fig.~\ref{fig:rshpn-three-fused-places-example}. In this net such a~place is indicated as $\pnFusionPlaceX{(\pnHierarchicalX{1}, \pnHierarchicalX{2}),(\alpha,\beta)}$ (from the perspective of net \pnHierarchicalX{1}) or $\pnFusionPlaceX{(\pnHierarchicalX{2},\pnHierarchicalX{1}),(\beta,\alpha)}$ (from the perspective of net \pnHierarchicalX{2}). More places can be fused, but not in RSHPN.
\begin{figure}
	\centering
	\includegraphics[width=0.8\linewidth]{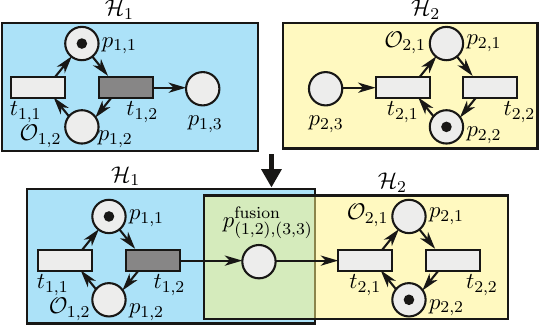}
	\caption{\mfb{Two places from two different nets fused with each other. When \pnTransitionX{1,2} fires then the token appears in \pnPlaceX{1,2} and in the
		fused place \pnFusionPlaceX{(1,2),(3,3)} (visible as \pnFusionPlaceX{(1,2),(3,3)} from perspective of \pnHierarchicalX{1} and as \pnFusionPlaceX{(2,1),(3,3)} from perspective of \pnHierarchicalX{2}). Whenever a~transition \pnTransitionX{2,1} fires the token disappears from the fused place.}
	}
	\label{fig:rshpn-three-fused-places-example}
\end{figure}
}  % OK

% !TeX root = ../figat_analiza.tex

\section{PN properties to be analysed}
\label{sec:pn-properties-to-be-analysed}

%%%%%%%%%%%%%%%%%%%%%%%%%%%%%%%%%%%%%%%%%%%%%%%%%%%%%%%%%%%%%%%
\subsection{Selected PN properties to be analysed}

{\bf PN is deadlock-free:}  if for any reachable marking $M_m$ (i.e.\ derived from $M_0$) there is always at least one fireable transition (i.e.\ enabled with a fulfilled condition). Absence of global deadlock in \pnHierarchicalX{} does not imply local deadlock cannot occur in any subsystem or between communicating subsystems. Therefore, individual lower level PNs need to be analysed.\\
{\bf PN is safe:} if for any reachable marking $M_m$  each place $p_i$ contains at the most one token, i.e. $0\leq M_m(p_i) \leq 1$. Safety assures that the operation associated with the place $p_i$ cannot be executed concurrently to itself.\\
{\bf PN is \bf conservative:}~\cite{Murata:1989:Petri:Nets} if  $\forall_{p_i\in\hat{p}} \; {\bf y}(p_i)>0$
% ($\bf y$ -- vector of weights for each place)
the weighted sum of tokens for every reachable marking $M_m$ is constant, i.e.: $\sum_{p_i\in\hat{p}} {\bf y}(p_i)\cdot M_m(p_i)=\sum_{p_i\in\hat{p}}^{} {\bf y}(p_i)\cdot M_0(p_i)= \rm const$.
If ${\bf y}(p_i)\geq0$ and ${\bf y}\neq{\bf 0}$ then PN is {\bf partially conservative}, while if ${\bf y}=(1,\ldots,1)^\top$ then PN is {\bf strictly conservative}, i.e.\ the number of tokens in PN remains the same for any $M_m$.

\subsection{\mfb{Justification of the selection of analysed properties}}
\label{subsec:justification-of-properties-analysis}
\mfb{
	{\bf Conservativeness analysis:}
	The structure of \pnHierarchicalX{} assures that when any subnet is activated, a~single token appears in its input place. Since each network in \pnHierarchicalX{} is meant to be safe, the parent PN cannot introduce another token to an already activated subnet. 	The number of tokens in an activated subnet changes only by firing its transitions. Hence, a~subnet is conservative and to what extent depends on its structure, not on the parent network. In effect the parent PN can be treated as a~transition connecting the output place to the input place of the analysed subnet. Hence, in a strictly conservative PN at any time the number of tokens is always constant and exactly one, i.e.\ subnet will always execute exactly one operation at any time (as it has a single token in an input place in the initial marking $M_0$). Ultimately, each subnet in an RSHPN ends up with exactly one token in its output place. Changes in PN marking can be examined through conservativeness analysis.
}

\mfb{
	{\bf Deadlocks analysis:}
	The interaction between subsystems can cause deadlocks, if they have inadequate structure or are inappropriately initialised. If two subsystems communicating bidirectionally both use the blocking mode the result depends on their proper initialisation. If both subsystems simultaneously start their operations by sending data to the other subsystem, then both will  start waiting for confirmation of data receipt, thus they will wait forever. 
However, if the two subsystems are properly initialised, i.e.\ one system starts by sending data while the other by waiting for this data to arrive, deadlock will not happen. Such situation should be detected through the formal analysis.
}

\mfb{	{\bf Safety analysis:}
	Ensuring the safety property in a PN guarantees that the number of tokens in each place (and also in a page) of the net is limited to a single token for any reachable marking. This means that there is no place in the net where, for any reachable marking, there is more than one token. This property is crucial in the context of RSHPN, where places are associated with operations. The presence of a token in a place signifies that the operation associated with that place is being executed. If this property were not satisfied, it would be possible to execute many operations in parallel using the same resources, e.g.\ simultaneously sending the same data to another subsystem -- this is not only incorrect but also unnecessary. \\
	Ensuring the safety property also guarantees that no more than a single token appears in a page at any given time. A single token in a page indicates that the lower-order net is active and executing its designated activity. Guaranteeing that no additional tokens can appear in the subnet while it is active ensures that, upon execution, no extra token will be introduced into its input place. This allows the analysis of the net to be conducted independently of the higher-layer net (the one containing a page with a single token). It is therefore possible to separate the analysis of the PN from the higher level network (as shown in Fig.~\ref{fig:rshpn_decomposition_2c}). Failure to maintain this property would render the proposed decomposition-based analysis method infeasible.
}

\subsection{Place and transition invariants}
\label{subsec:place_transition_invariant_analysis}
%%%%%%%%%%%%%%%%%%%%%%%%%%%%%%%%%%%%%%%%%%%%

Invariants~\cite{Reisig:2013:place_invariants,Murata:1989:Petri:Nets} are properties of the logical structure of a~network. They characterize the way transitions are fired~\cite{Freedman:1991}.

{\bf Place invariants} express properties of reachable markings~\cite{Reisig:2013:place_invariants}, i.e.\ they describe sets of places in PN where the weighted sum of tokens for each reachable marking remains constant.
Let $\bf y$ be the vector of weights assigned to places for which the assumption of place invariants is satisfied. This is equivalent to:
$		{\begin{array}{c}
		M_{m}^\top \cdot {\bf y} = M_{0}^\top \cdot {\bf y} = \rm const,
	\end{array}
}
$
i.e.: $(M_{m} - M_{0})^\top \cdot {\bf y}=\bf 0$, and by using Eq.~(\ref{eq:reachable_marking_from_initial_marking_5}) we obtain:
\begin{equation}
	{\begin{array}{c}
			{\bf x}^\top \cdot (N \cdot {\bf y}) = {\bf 0},\ \textrm{and since}\ {\bf x} \neq {\bf0}\ \textrm{then:}\ N \cdot {\bf y} = \bf 0
		\end{array}
	}
	\label{eq:p_invariant_3}
\end{equation}
The solutions of Eq.~(\ref{eq:p_invariant_3}) for $\bf y\neq \bf 0$ are called place invariants (guaranteeing constant weighted sum of tokens).
The set of places corresponding to non-zero elements of the solution is $\left\Vert \bf y \right\Vert$, called the support of place invariant~\cite{Murata:1989:Petri:Nets}.
Firing any transition in PN does not change the weighted sum of tokens from $\left\Vert \bf y \right\Vert$. This enables  distinction of the set of places representing operations repeated cyclically and those dependent on each other. For example, if the sum of tokens is 1, then only one operation is executed from those represented by the places in $\bf ||y||$. By obtaining two independent vectors {$\bf y_1$} and {$\bf y_2$}, one can observe parallel operations, such as those performed by two simultaneously operating subsystems.

%%%%%%%%%%%%%%%%%%%%
%%%%%%%%%%%%%%%%%%%%
{\bf Transition invariants} are vectors defining the number of transition firings keeping the marking of PN the same, i.e.\ after firing transitions as many times as specified in the transition invariant vector, PN will switch from $M_0$ to $M_m$, but $M_m = M_{0}$. Hence, based on Eq.~(\ref{eq:reachable_marking_from_initial_marking_5}) we obtain $N^{\top} \cdot \bf x = \bf 0$. 
The non-zero solutions of this equation
are transition invariants.
The set of transitions corresponding to the non-zero elements of the solution is $\left\Vert \bf x \right\Vert$, called the support of transition invariant~\cite{Murata:1989:Petri:Nets}, i.e.\ firing all transitions from $\left\Vert \bf x \right\Vert$ does not change PN marking, e.g.\ if ${\bf x} = (1,2)^\top$ and PN starts from $M_0$ it will reach $M_0$ by firing: $t_1$ once and $t_2$ twice, i.e. $M_0 \xrightarrow{(t_1), (t_2)^{2}} M_0$.
The transition invariant does not specify the order of transition firings.
Transition invariants can be used to prove the repeatability of the system and the absence of deadlocks while firing transitions from $\bf ||x||$.

%%%%%%%%%%%%%%%%%%%%%%%%%%%%%%%%%%%%%%%%%%%%%%%%%%%%%%%%%%%%%%%%%%%%%%%%%%%%%%%%
%%%%%%%%%%%%%%%%%%%%%%%%%%%%%%%%%%%%%%%%%%%%%%%%%%%%%%%%%%%%%%%%%%%%%%%%%%%%%%%%

 % OK
% !TeX root = ../figat_analiza.tex

\section{HPN analysis methods}
\label{sec:pn-analysis-methods}

\subsection{HPN analysis decomposition}
\label{subsec:rshpn-decomposition}

\begin{figure}
	\begin{subfigure}{0.49\linewidth}
		\centering
		\includegraphics[width=0.95\linewidth]{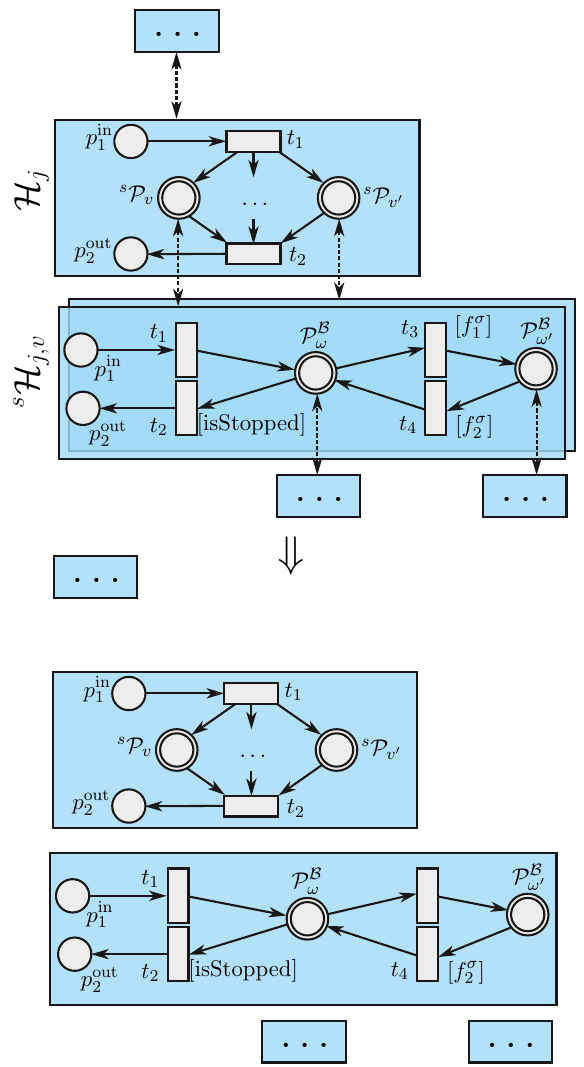}
		\caption{}
		\label{fig:rshpn_decomposition_2d}
	\end{subfigure}
	\begin{subfigure}{0.49\linewidth}
		\centering
		\includegraphics[width=0.7\linewidth]{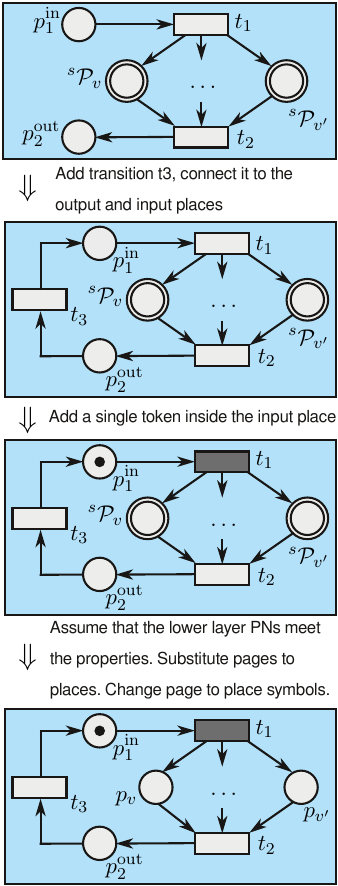}
		\caption{}
		\label{fig:rshpn_decomposition_2c}
	\end{subfigure}
	\caption{\mfb{Method of isolating a PN: (a) Removing connections to higher layers; (b) The first two transformations involve adding an additional transition, connecting it to input and output places, and introducing a token; final transformation replaces pages with places, ensuring isolation from lower layers.}}
	\label{fig:rshpn_decomposition_2d_rshpn_decomposition_2c}
\end{figure}

\mfb{The proposed method decomposes RSHPN analysis into analysing individual subnets, significantly mitigating the state explosion issue in reachability graphs. This decomposition involves isolating each network from higher and lower layers:
	\begin{itemize}
		\item \textbf{Isolation from higher layers:} This requires the addition of an extra transition connecting the output and input places of the analysed net. This ensures that when a token appears at the output place, the execution of the analysed PN is not terminated, but a new iteration of execution of this PN is started (firing the newly added transition instead of a return of the token to the higher layer PN causes this token to return to the input place of the analysed net). This ensures that properties can be analysed independently of higher-layer networks.
		\item \textbf{Isolation from lower layers:} This requires the replacement of pages with places, only for the analysis purposes. The verified properties of the modified network remain valid for the hierarchical network if the same properties hold for lower layer networks. Consequently, verifying properties in an isolated network requires ensuring that these properties also hold in its subordinate networks.
\end{itemize}}
\mfb{Fig.~\ref{fig:rshpn_decomposition_2d_rshpn_decomposition_2c} presents the isolation method. It follows the reduction methods outlined in~\cite{Murata:1989:Petri:Nets}. Instead of analysing a highly complex network, decomposition  enables the study of many simpler disjoint subnets preventing state explosion.}

\subsection{Methods of HPN analysis}
\label{subsec:rshpn-model-analysis-introduction}

%%%%%

\mfb{Once the RSHPN is decomposed into separate PNs, the analysis of each PN can be performed using one of three methods: 1) PN graphical interpretation, 2) reachability graph construction, or 3) place/transition invariants analysis. The choice of method depends on the complexity of the network being analysed. A discussion of network complexity is presented in Section~\ref{subsec:pn-complexity-assessment}, while guidelines for selecting the appropriate analysis method for a given network are provided in Section~\ref{subsec:decision-on-pn-analysis-method-selection}.}

{\bf PN interpretation:}
It consists in firing enabled transitions and presenting graphically all possible markings $M_m$ that can be derived from $M_0$, as shown in Fig.~\ref{fig:analysismultiagentroboticsystemlayer} and Fig.~\ref{fig:analysismultiagentroboticsystemlayer_simplified}.

{\bf Reachability graph analysis:}
Reachability graph nodes represent markings $M_m$ derived from $M_0$, while edges show transformations of markings due to firing transitions $t$, e.g.\  $M_0 \xrightarrow{} \ldots \xrightarrow{} M_m \xrightarrow{t} M_{m+1}$.
The analysis consists in deriving from $M_0$ all possible markings (e.g.\ Fig.~\ref{tab:behaviour_reachability_tree_analysis}) and checking that for each marking:
(1) there is always at least one transition that can fire, i.e. PN has no deadlock;
(2) number of tokens in each place is either 0 or 1, i.e.\ PN is safe; and
(3) total number of tokens in any marking is same (strictly conservative) or variable (partially conservative).

{\bf Place/transition invariants analysis:}
\label{subsubsec:place_transition_invariant_analysis}
%%%%%%%%%%%%%%%%%%%%%%%%%%%%%%%%%%%%%%%%%%%%
%%%%%%%%%%%%%%%%%%%%%%%%%%%%%%%%%%%%%%%%%%%%
Place invariants are used to study two properties: 1)~network safety, and 2)~network conservativeness. Transition invariants show the lack of deadlocks in the network. Safety analysis requires solving Eq.~(\ref{eq:p_invariant_3}). If all places of the network belong to $\left\Vert \bf y \right\Vert$ and $M_0$ is bounded (i.e., the sum of tokens in $M_0$ is finite) then the network is bounded. If for each $\left\Vert \bf y \right\Vert$ the sum of tokens is 1 then the network is safe.

Conservatism analysis also employs solutions of Eq.~(\ref{eq:p_invariant_3}). The vector of weights can be obtained just by summing all the linearly independent vectors that are place invariants. If the resulting vector is $(1,\ldots,1)^{T}$, then the network is strictly conservative. If no place invariant is found for the network except
$\bf y = \bf 0$, then the network is not conservative. On the other hand, if some component of the vector is 0 (but $\bf y\neq \bf 0$) then the network is partially conservative. In other cases, the network is conservative.
%%%%%%%%%%%%%%%%%%%%%%%%%%%%%%%%%%%%%%%%%%%%

%%%%%%%%%%%%%%%%%%%%%%%%%%%%%%%%%%%%%%%%%%%%

A network is free of deadlocks if it is live~\cite{Murata:1989:Petri:Nets}. A network is considered live if, for every reachable marking $M_m$, there exists a sequence of transition firings in which each transition can potentially fire at least once. If, based on $N^{\top} \cdot \bf x = \bf 0$,
we can find a vector $\bf{x}$ that is a transition invariant with non-zero values for each transition, then firing all transitions indicated in $\mathbf{x}$ from any reachable marking $M_m$ will return the marking to $M_m$. According to~\cite{Murata:1989:Petri:Nets}, this confirms that the network is live. Thus, if every transition can potentially fire from each reachable marking $M_m$, there are no deadlocks in the network.

 % OK
% !TeX root = ../figat_analiza.tex

\section{Robotic System Hierarchical Petri Net}
\label{sec:rshpn_metamodel}
\begin{figure}% figure* - for two column view
	\centering
	\includegraphics[width=0.92\linewidth]{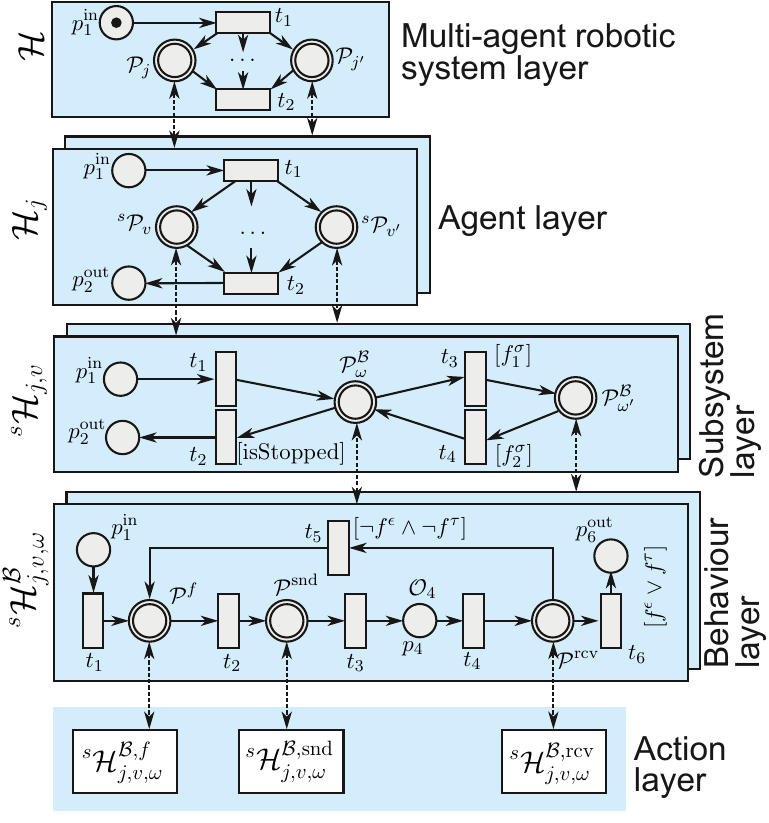} %hpn_nowa_uproszczona % hpn_six_layers_new2
	\caption{RSHPN \pnHierarchicalXXX{}{}{} modeling activities of a robotic system \mfb{(adapted from~\cite{Figat:2022:RAL})}}
	\label{fig:hpnnowauproszczona}
\end{figure}

RSHPN \pnHierarchicalXXX{}{}{} is a HPN adopted to model the activity of the whole multi-agent robotic system \roboticSystem{}. It consists of six layers (Fig.~\ref{fig:hpnnowauproszczona}).
RSHPN decomposition into layers reflects the structure of \roboticSystem~\cite{Figat:2022:RAS}.
Each RSHPN layer describes system activities at a different  level of abstraction: 1) system composed of agents, 2) agent structure, 3) task definition for subsystems, 4) behaviour definition for each subsystem (a pattern parametrised by a transition function), 5) decomposition of a transition function and communication model definition, 6) general parametrised model of communication. This defines the robotic system's structure and the activity of its components. The initial marking is in Fig.~\ref{fig:hpnnowauproszczona}. A single token enters at the top and propagates through lower layers, multiplying appropriately as it moves.

\mfb{A detailed description of individual elements in the RSHPN meta-model is provided in~\cite{Figat:2022:RAS}, while an even more comprehensive explanation in the supplementary video  (\textit{Robotic System Hierarchical Petri Net}\footnote{\mfb{Chapter 4: \textit{Robotic System Hierarchical Petri Net}, available at \url{https://www.youtube.com/watch?v=JenAB1IKVVY&t=2295s}.}} and \textit{How RSHPN Works}\footnote{\mfb{Chapter 5: \textit{How RSHPN Works}, available at \url{https://www.youtube.com/watch?v=JenAB1IKVVY&t=3434s}.}}).}

%%%%%%%%%%%%%%%%%%%%%%%%%%%%%%%%%%%%%%%%%%%%%%%%%%%%%%%%%%%%%
%%%%%%%%%%%%%%%%%%%%%%%%%%%%%%%%%%%%%%%%%%%%%%%%%%%%%%%%%%%%%

%%%%%%%%%%%%%%%%%%%%%%%%%%%%%%%%%%%%%%%%%%%%%%%%%%%%%%%%%%%%%
%%%%%%%%%%%%%%%%%%%%%%%%%%%%%%%%%%%%%%%%%%%%%%%%%%%%%%%%%%%%%

\subsection{Description of RSHPN layers}
\label{subsec:description-of-rshpn-layers}

{\bf Multi-agent robotic system layer:} contains a~single net \pnHierarchicalXXX{}{}{} (Fig.~\ref{fig:hpnnowauproszczona}) representing the \roboticSystem\ task \subsystemTask\ aggregating tasks executed by all agents (activities of \agentX{j} are represented by \pnPageXXX{j}{}{}). At system initiation \pnTransitionX{1} fires, activating in parallel all pages for \roboticSystem\ agents. Once all pages complete their activities, \pnTransitionX{2} fires and the system terminates its activities.

{\bf Agent layer:} defines a single \pnHierarchicalXXX{j}{}{} for each $\agentX{j}$. \pnHierarchicalXXX{j}{}{} represents its task \subsystemTaskX{j} aggregating the tasks of its subsystems. \pnHierarchicalXXX{j}{}{} contains a page \pnPageXXX{j,v}{}{s} for each \subsystemX{j,v} in \agentX{j}~\cite{Figat:2022:RAS,Figat:2022:RAL}. All pages activate in parallel, allowing subsystems to act independently.

{\bf Subsystem layer:} defines for each $\subsystemX{j,v}$ a~single net \pnHierarchicalXXX{j,v}{}{\subsystemSymbol} representing the task \subsystemTaskXX{j,v}{s}. The task determines how $\subsystemX{j,v}$ switches between its behaviours  \behaviourX{j, v, \omega}. Each \behaviourX{j, v, \omega} is represented by a page \pnPageXBS{j,v,\omega}{}. For each pair of consecutive behaviors (e.g.\ \behaviourX{j, v, \omega} -- current, and \behaviourX{j, v, \omega'} -- next behaviour), there is a~transition \pnTransitionXXX{j, v, \alpha}{}{s} with initial condition \initialConditionX{j,v,\alpha}. When  \behaviourX{j, v, \omega} terminates and \initialConditionX{j,v,\alpha} is true, it switches to \Bsbbj{j, v, \omega'}. A single condition associated with transitions \pnTransitionXXX{j, v, \alpha}{}{s} from \pnPageXXS{j,v,\omega}{\agentBehaviour} must be true at termination of \behaviourX{j, v, \omega}. Condition $\rm [isStopped]$ is true if \subsystemX{j,v} terminates.

{\bf Behaviour layer:} defines for each behaviour \behaviourX{j,v,\omega} a~single fixed-structure net \pnHierarchicalXXX{j,v,\omega}{\behaviourSymbol}{\subsystemSymbol} composed of pages for an elementary action \elementaryActionX{j,v,\omega}, with terminal \terminalConditionX{j,v,\omega} and error \errorConditionX{j,v,\omega} conditions.  \elementaryActionX{j,v,\omega} consists of 3 pages and one operation, all executed unconditionally: 1) \pnPageXBS{j,v,\omega}{\transitionFunction} calculates the transition function \transitionFunctionX{j,v,\omega}, 2) \pnPageXBS{j,v,\omega}{\send} sends out the results of \transitionFunctionX{j,v,\omega} evaluation, 3) increments the discrete time stamp $\iota$ (operation \pnOperationXBS{j,v,\omega,4}{}), 4) \pnPageXBS{j,v,\omega}{\receive} inserts the received data  into \inputBufferXX{j,v}{}. If neither \errorConditionX{j,v,\omega} nor terminal \terminalConditionX{j,v,\omega} are satisfied, \pnTransitionXBS{j, v, \omega, 5}{} fires leading to the next iteration of \behaviourX{j, v, \omega}. Otherwise,  \pnTransitionXBS{j, v, \omega, 6}{} fires, terminating \behaviourX{j, v, \omega}, i.e.\ a~single token appears in \pnPlaceXBS{j,v,\omega,6}{} and the control returns to \pnHierarchicalXXS{j,v}{} which designates a~new behaviour based on \subsystemTaskXS{j,v}.

{\bf Action layer:} defines 3 modules each divided into 2 layers for each behaviour $\behaviourX{j,v,\omega}$:
1) \pnHierarchicalXXX{j,v,\omega}{\behaviourSymbol,\transitionFunction}{s} -- defines transition function \transitionFunctionX{j,v,\omega} executed by \subsystemX{j,v}  realising \behaviourX{j, v, \omega},
2)~\pnHierarchicalXXX{j,v,\omega}{\behaviourSymbol,\send}{\subsystemSymbol} represents the send communication mode used by \subsystemX{j,v},
3) \pnHierarchicalXXX{j,v,\omega}{\behaviourSymbol,\receive}{\subsystemSymbol} represents the receive communication mode used by \subsystemX{j,v}.
For a more detailed description see~\cite{Figat:2022:RAS}.

\subsection{Communication models}
\label{subsec:rshpn-communication-models}

\begin{figure}
	\centering
	\includegraphics[width=1.0\linewidth]{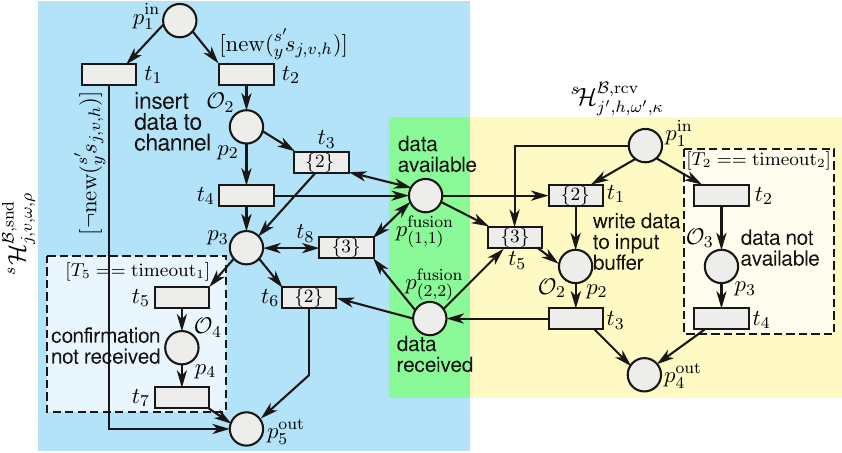}
	\caption{General communication model for subsystems: $s_{j,v}$ (sender) and $s_{j',h}$ (receiver); 9 possible models \mfb{(adapted from~\cite{Figat:2022:RAS})}.}
	\label{fig:general_communication_modes_petri_net}
\end{figure}

Subsystem \subsystemX{j,v} communication modes are defined by two nets: a) \pnHierarchicalXXX{j,v,\omega}{\behaviourSymbol,\send}{\subsystemSymbol} and b) \pnHierarchicalXXX{j,v,\omega}{\behaviourSymbol,\receive}{\subsystemSymbol}. Each determines  the order in which the \subsystemX{j,v} communicates with its associated subsystems (the arrangement of pages). For each communicating pair of subsystems an inter-subsystem communication model is produced from these modes (Table~\ref{tab:communication_mode_pattern_table}). Since each subsystem can communicate in three modes: blocking, blocking with timeout or non-blocking, nine possible inter-subsystem communication models result. All are realised by the PN presented in Fig.~\ref{fig:general_communication_modes_petri_net}.

\begin{table}[htb]
	\caption{Nine communication models realised by the PN in Fig.~\ref{fig:general_communication_modes_petri_net} depending on  $\rm t\_out_1$ and $\rm t\_out_2$, where ($\rm t\_out_{\gamma} \triangleq timeout_{\gamma}, \gamma=1,2$);  modes: $\rm NB$ -- non-blocking, $\rm B$ -- blocking, and $\rm BT$ -- blocking with timeout}
	\scalebox{0.95}{
	\scriptsize
	{\renewcommand{\arraystretch}{1.0}
		\begin{tabular}{|cc||c|c|c|}
			\hline
			\multicolumn{2}{|c||}{\multirow{1}{*}{Communication}} & \multicolumn{3}{c|}{Receive} \\ \cline{3-5}
			\multicolumn{2}{|c||}{mode} & $\rm t\_out_2=0$ & $\rm t\_out_2=\infty$ & $0<\rm t\_out_2<\infty$\\ \hline \hline
			\multicolumn{1}{|c|}{\multirow{3}{*}{\rotatebox[origin=c]{90}{   Send   }}} & $\rm t\_out_1=0$ &
			% The first row with modes combinations
			NB-NB & NB-B & NB-BT \\ \cline{2-5}
			\multicolumn{1}{|c|}{}& $\rm t\_out_1=\infty$ &
			% The second row with modes combinations
			B-NB & B-B & B-BT \\ \cline{2-5}
			\multicolumn{1}{|c|}{}& $0<\rm t\_out_1<\infty$ &
			% The third row with modes combinations
			BT-NB & BT-B & BT-BT \\ \hline
	\end{tabular}}
	}
	\label{tab:communication_mode_pattern_table}
\end{table}

\subsection{Complexity of RSHPN layers}
\label{subsec:discussion_rshpn_complexity}
RSHPN \pnHierarchicalX{} consists of four network types:
\begin{enumerate}
	\item trivial, variable structure: \pnHierarchicalX{}, \pnHierarchicalX{j}, and assuming sequential or parallel arrangement: a)~\pnHierarchicalXXX{j,v,\omega}{\mathcal{B},f}{s}, b)~\pnHierarchicalXXX{j,v,\omega}{\mathcal{B},\rm snd}{s} and c)~\pnHierarchicalXXX{j,v,\omega}{\mathcal{B},\rm rcv}{s}, and \pnHierarchicalXXX{j,v,\omega,\psi}{\mathcal{B},f}{s},
	\item trivial, fixed structure: \pnHierarchicalXXX{j,v,\omega}{\mathcal{B}}{s},
	\item complex, fixed structure: a~network composed of \pnHierarchicalXXX{j,v,\omega, \rho}{\mathcal{B},\rm snd}{s} and \pnHierarchicalXXX{j,v,\omega,\kappa}{\mathcal{B},\rm rcv}{s} representing the general communication model for a~pair of subsystems,
	\item complex, variable structure: \pnHierarchicalXXX{j,v}{}{s}, and in hybrid arrangements: \pnHierarchicalXXX{j,v,\omega}{\mathcal{B},f}{s}, \pnHierarchicalXXX{j,v,\omega}{\mathcal{B},\rm snd}{s}, \pnHierarchicalXXX{j,v,\omega}{\mathcal{B},\rm rcv}{s}.
\end{enumerate}
PN structure complexity affects the choice of analysis method: for simple PNs, use PN interpretation and reachability graphs, while for complex PNs, use place and transition invariants.
 % OK

% !TeX root = ../figat_analiza.tex

\section{RSHPN meta-model analysis}
\label{sec:rshpn_model_analysis}
PNs in RSHPN layers are analysed from the top layer to lower layers, as shown in the video tutorial\footnote{Introduction to the RSHPN analysis: https://youtu.be/JenAB1IKVVY}.

\subsection{Multi-agent robotic system layer analysis}
\label{subsec:rshpn-analysis-multi-agent-robotic-system-layer-analysis}
\begin{figure}
	\centering
	\includegraphics[width=1.0\linewidth]{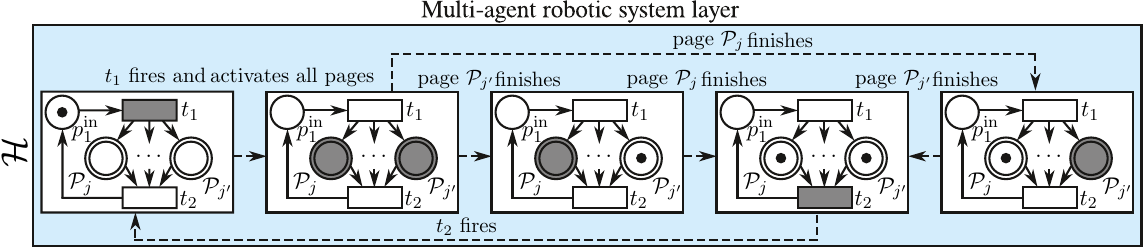}
	\caption{
		Graphical analysis of \pnHierarchicalX{} (first layer of RSHPN in Fig.~\ref{fig:hpnnowauproszczona})
}
	\label{fig:analysismultiagentroboticsystemlayer}
\end{figure}

\begin{figure}
	\centering
	\includegraphics[width=0.7\linewidth]{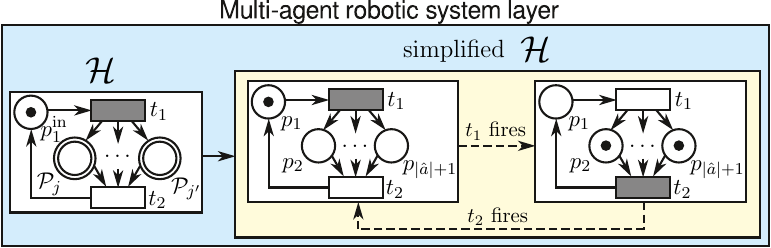}
	\caption{
		Simplified \pnHierarchicalX{} has two markings: $M_0=(1,0,\ldots,0)^\top$ and $M_1=(0,1,\ldots,1)^\top$; $|\hat{a}|$ -- number of agents in \roboticSystem
	}
	\label{fig:analysismultiagentroboticsystemlayer_simplified}
\end{figure}

%%%%%%%%%%%%%%%%%%%%%%%%%%%%%%%%%%%%%%
%%%%%%%%%%%%%%%%%%%%%%%%%%%%%%%%%%%%%%
\pnHierarchicalX{} is composed of one PN. Its simplified structure is shown in Fig.~\ref{fig:analysismultiagentroboticsystemlayer}. Assuming that lower-level PNs are correct the  simplified PN is as in Fig.~\ref{fig:analysismultiagentroboticsystemlayer_simplified}. By inspection we conclude that \pnHierarchicalX{} is safe (in every reachable marking, i.e. $M_0$ and $M_1$, each place has at the most a single token), conservative (due to the vector of weights ${\bf y}=(|\hat{a}|,1,\ldots,1)^\top$) and deadlock-free as in $M_0$ and $M_1$, always a single transition can fire.

%%%%%%%%%%%%%%%%%%%%%%%%%%%%%%%%%%%%%%%%%
%%%%%%%%%%%%%%%%%%%%%%%%%%%%%%%%%%%%%%%%%
%%%%%%%%%%%%%%%%%%%%%%%%%%%%%%%%%%%%%%%%%

\subsection{Agent layer analysis}
\label{subsec:rshpn-analysis-agent-layer-analysis}

\begin{figure}
	\includegraphics[width=1.0\linewidth]{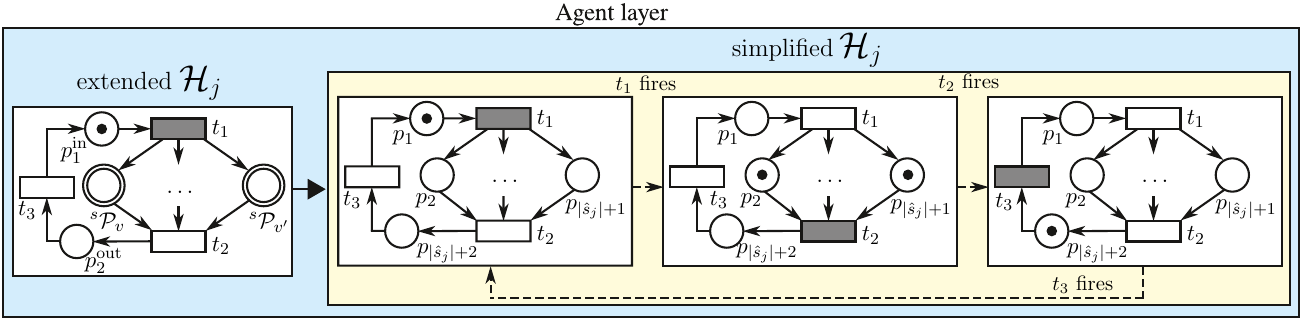}
	\caption{Simplified \pnHierarchicalX{j}; $|\hat{s}_j|$ -- number of subsystems in \agentX{j}; \footnotesize{$M_0=(1,0,\ldots,0,0)^\top$, $M_1=(0,1,\ldots,1,0)^\top$, $M_2=(0,0,\ldots,0,1)^\top$}}
	\label{fig:analysisagentlayernew_simplified}
\end{figure}

%{\bf Agent layer analysis:}
The agent layer consists of \setSize{\agentSet} different \pnHierarchicalX{j}. Each \pnHierarchicalX{j} consists of a single input and single output place. Thus, the structure of \pnHierarchicalX{j} can be extended with an additional transition.
Assuming that the properties of lower level PNs, i.e. \pnHierarchicalX{j,v}, are fulfilled, then the extended \pnHierarchicalX{j} can be further simplified by changing pages into places, as in Fig.~\ref{fig:analysisagentlayernew_simplified}. Thus, the analysis of \pnHierarchicalX{j} boils down to the analysis of the simplified PN.
The number of tokens in each place within each reachable marking (i.e. $M_0$, $M_1$, $M_2$) is limited to 1, so \pnHierarchicalX{j} is safe. Since from any reachable marking there is always a single transition ready to fire, the \pnHierarchicalX{j} is deadlock-free.
A vector of weights ${\bf y}=(|\hat{s}_j|, 1, \ldots, 1, |\hat{s}_j|)^\top$ ensures that in the simplified \pnHierarchicalX{j} the weighted sum of tokens is equal to $|\hat{s}_j|$, which means that \pnHierarchicalXXX{j}{}{} is conservative.

\subsection{Subsystem layer analysis}
\label{subsec:rshpn-analysis-subsystem-layer-analysis}

\begin{figure}
	\centering
	\includegraphics[width=0.95\linewidth]{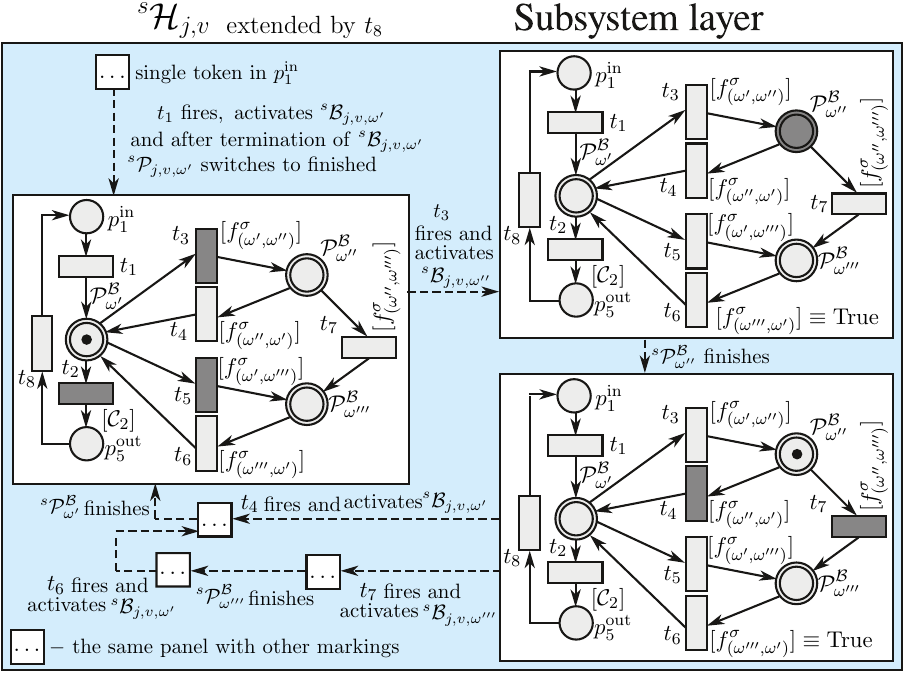}
	\caption{Analysis of an exemplary \pnHierarchicalXXX{j,v}{}{s}
		}
	\label{fig:analysis_subsystem_layer}
\end{figure}

Subsystem layer contains $\sum_{\agentX{j}\in\agentSet} \setSize{\subsystemSetX{j}}$ panels, each holding a~single PN \pnHierarchicalXXX{j,v}{}{s}. Analysis of \pnHierarchicalXXX{j,v}{}{s} requires substitution of \pnHierarchicalX{j} by a transition connecting \pnHierarchicalXXX{j,v}{}{s} output with input place.
Since \pnHierarchicalXXX{j,v}{}{s} represents an arbitrary task \subsystemTaskXX{j,v}{s} it assumes diverse structures.
Hence, the analysis of \pnHierarchicalXXX{j,v}{}{s} is based on an incidence matrix. A valid structure of \pnHierarchicalXXX{j,v}{}{s} requires no self-loops and that within each row of incidence matrix $^{s}N_{j,v}$ exactly two non-zero numbers exist: $-1$ and $1$. If the assumptions are violated the PN structure is incorrect.
E.g.\ for the PN in Fig.~\ref{fig:analysis_subsystem_layer} the incidence matrix $^{s}N_{j,v}$ is as in Eq.~(\ref{eq:subsystem-layer-analysis-petri-net-matrix_first_property}).
 \begin{equation}
	\centering
	{\small ^{s}N_{j,v} =
		\begin{blockarray}{c@{}c@{}c@{}c@{}cc}
			p_1^{\rm in} & ^{s}\mathcal{P}_{j,v,\omega'}^{\mathcal{B}} & ^{s}\mathcal{P}_{j,v,\omega''}^{\mathcal{B}} & ^{s}\mathcal{P}_{j,v,\omega'''}^{\mathcal{B}} & p_5^{\rm out}\\
			\begin{block}{(c@{}c@{}c@{}c@{}c)c}
				-1&1&0&0&0&\pnTransitionX{1}\\
				0&-1&0&0&1&\pnTransitionX{2}\\			
				0&-1&1&0&0&\pnTransitionX{3}\\
				0&1&-1&0&0&\pnTransitionX{4}\\
				0&-1&0&1&0&\pnTransitionX{5}\\									
				0&1&0&-1&0&\pnTransitionX{6}\\
				0&0&-1&1&0&\pnTransitionX{7}\\
				1&0&0&0&-1&\pnTransitionX{8}\\									
			\end{block}
		\end{blockarray}
	}
	\label{eq:subsystem-layer-analysis-petri-net-matrix_first_property}
\end{equation}
Satisfaction of the above requirements ensures strict conservativeness. To show the invariability of the token number in any $M_m$, it suffices to notice that when a transition fires, one token from a place/page is consumed, and one token is inserted into another place/page, hence $y=(1,\ldots,1)^\top$.

%%%%%%%%%%%%%%%%%%%%%%%%%%%%%%%%%%%%
%%%%%%%%%%%%%%%%%%%%%%%%%%%%%%%%%%%%

%%%%%%%%%%%%%%%%%%%%%%%%%%%%%%%%%%
%%%%%%%%%%%%%%%%%%%%%%%%%%%%%%%%%%
%%%%%%%%%%%%%%%%%%%%%%%%%%%%%%%%%%
%%%%%%%%%%%%%%%%%%%%%%%%%%%%%%%%%%%%%%%%%%
%%%%%%%%%%%%%%%%%%%%%%%%%%%%%%%%%%%%%%%%%%

To avoid deadlock, in any $M_m$ exactly one transition should be able to fire, i.e.\ the condition associated with the transition must be met. To ensure deterministic execution of \pnHierarchicalXXX{j,v}{}{s} conditions associated with transitions for which considered place/page is an input must satisfy:
	(1)~the logical sum of all initial conditions must be $\rm True$;
	and
	(2)~the logical product of any two conditions must be $\rm False$ (e.g.
		if a token is in \pnPageXXX{j,v,\omega'}{\mathcal{B}}{s} (Fig.~\ref{fig:analysis_subsystem_layer}), then: ${\footnotesize\big[\initialConditionX{(\omega',\omega'')} \lor\  \initialConditionX{(\omega',\omega''')} \lor\ \pnConditionX{2}\big] =\text{\rm True}}$, and ${\footnotesize\begin{array}{c}\big[(\ \initialConditionX{(\omega',\omega'')} \land\  \initialConditionX{(\omega',\omega''')} ={\rm False}\ )\ \land (\ \initialConditionX{(\omega',\omega'')} \land\  \pnConditionX{2} ={\rm False}\ )\land\end{array}}$ ${\footnotesize \begin{array}{c}\land (\ \initialConditionX{(\omega',\omega''')} \land\ \pnConditionX{2} ={\rm False}\ )\big]={\rm True}\end{array}}$).
 Both conditions must be met at the end of place/page activity.
The first condition ensures at least one transition can fire, preventing deadlock in \pnHierarchicalXXX{j,v}{}{s}. The second condition ensures only one behaviour can be selected, ensuring determinism.
%
%%%%%%%%%%%%%%%%%%%%%%%%%%%%%%%%%%%%%%%%%%
%%%%%%%%%%%%%%%%%%%%%%%%%%%%%%%%%%%%%%%%%%
%%%%%%%%%%%%%%%%%%%%%%%%%%%%%%%%%%%%%%%%%%

\subsection{Behaviour layer analysis}
\label{subsec:rshpn-analysis-behaviour-layer-analysis}
\begin{figure}

	\begin{subfigure}{0.49\textwidth}
		\centering
		\includegraphics[width=0.8\linewidth]{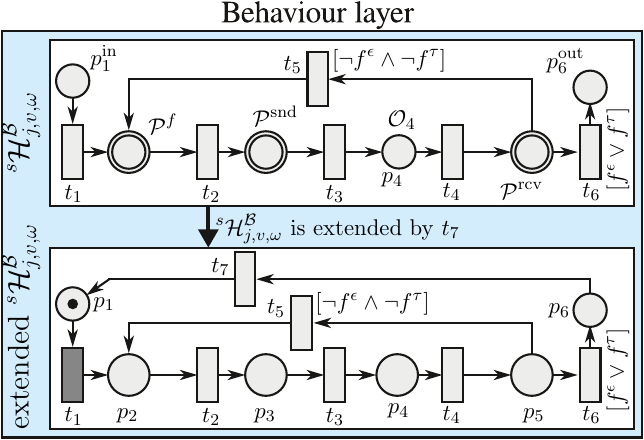}
		\caption{}
		\label{fig:analysis_behaviour_layer}
	\end{subfigure}
	\begin{subfigure}{0.49\textwidth}
		\centering
		\setlength{\tabcolsep}{2pt} % space between columns
		\renewcommand{\arraystretch}{1.4}
		\scalebox{0.7}{	
			\begin{tabular}{|c||c|c|c|c|c|c|c|c|}
				\hline
				\multicolumn{1}{|c||}{Marking} & \multicolumn{7}{c|}{Transitions}\\  \hline
				$M_{m}^{\mathcal{B}}$ =$\big(\mu^{m}(p_{1})$, \ldots, $\mu^{m}(p_{6})\big)^\top$&  $t_{1}$ & $t_{2}$& $t_{3}$& $t_{4}$& $t_{5}$& $t_{6}$& $t_{7}$\\ \hline \hline
				$M_{0}^{\mathcal{B}}=(1,0,0,0,0,0)^\top$ & $M_{1}^{\mathcal{B}}$ & - & - & - & - & - & - \\ \hline			
				$M_{1}^{\mathcal{B}}=(0,1,0,0,0,0)^\top$ & - & $M_{2}^{\mathcal{B}}$ & - & - & - & - & - \\ \hline
				$M_{2}^{\mathcal{B}}=(0,0,1,0,0,0)^\top$ & - & - & $M_{3}^{\mathcal{B}}$ & - & - & - & - \\ \hline
				$M_{3}^{\mathcal{B}}=(0,0,0,1,0,0)^\top$ & - & - & - & $M_{4}^{\mathcal{B}}$ & - & - & - \\ \hline
				$M_{4}^{\mathcal{B}}=(0,0,0,0,1,0)^\top$ & - & - & - & - & $M_{1}^{\mathcal{B}}$ & $M_{5}^{\mathcal{B}}$ & - \\ \hline
				$M_{5}^{\mathcal{B}}=(0,0,0,0,0,1)^\top$ & - & - & - & - & - & - & $M_{0}^{\mathcal{B}}$ \\ \hline
			\end{tabular}
		}
		\caption{}
		\label{tab:behaviour_reachability_tree_analysis}	
	\end{subfigure}
	\caption{(\ref{fig:analysis_behaviour_layer}): extended \pnHierarchicalXXX{j,v,\omega}{\mathcal{B}}{s};
		(\ref{tab:behaviour_reachability_tree_analysis}):
		tabular form of the reachability graph of extended \pnHierarchicalXXX{j,v,\omega}{\mathcal{B}}{s}
		(rows represent graph nodes, columns transitions (e.g.\ firing $t_4$ at $M_{3}^{\mathcal{B}}$, leads to $M_{4}^{\mathcal{B}}$)
	}
	\label{fig:analysis_behaviour_layer_pattern}
\end{figure}

%%%%%%%%%%%%%%%
%{\bf Behaviour layer analysis}
The behaviour layer consists of $\sum_{\agentX{j}\in\agentSet} \sum_{\subsystemX{j,v}\in\subsystemSetX{j}} \setSize{\behaviourSetX{j,v}}$ panels.
Each PN (panel) has exactly the same structure as \pnHierarchicalXXX{j,v,\omega}{\mathcal{B}}{s} -- this is the general behaviour \behaviourX{j,v,\omega} pattern (Fig.\ref{fig:hpnnowauproszczona}).
So instead of analysing all the PNs specified in all panels, it suffices to analyse a single PN being the pattern.
Hence, PN in Fig.~\ref{fig:analysis_behaviour_layer} is analysed only.
In that network each transition has exactly one input place and exactly one output place, i.e.\ firing any transition does not change the number of tokens in the network. Thus, the network is strictly conservative. Since there is exactly one token in the initial marking, the network is safe. Moreover, for any reachable marking, there is always exactly one fireable transition (the conditions for the transitions \pnTransitionX{5} and \pnTransitionX{6} are mutually exclusive), thus there is no deadlock.
Those properties can also be verified using the reachability graph presented in Table~\ref{tab:behaviour_reachability_tree_analysis}. For each possible marking: a) each place has at the most a~single token, i.e.\ PN is safe, b) total number of tokens remains always the same, i.e.\ the net is strictly conservative, and c) for any reachable marking always a~single transition can fire, i.e.\ net is deadlock-free.

%%%%%%%%%%%%%%%%%%%

%%%%%%%%%%%%%%%%%%%%%%%%%%
%%%%%%%%%%%%%%%%%%%%%%%%%%
%%%%%%%%%%%%%%%%%%%%%%%%%%%

%%%%%%%%%%%%%%%%%%%%%%%%%%%%%%%%%%%%%%
\subsection{Activity sublayers analysis}
\label{subsec:rshpn-analysis-activity-sublayers-analysis}
Analysis of the action layer requires the analysis of 3 PNs in \pnHierarchicalXXX{j,v,\omega}{\mathcal{B}}{s}, namely:
\pnHierarchicalXXX{j,v,\omega}{\mathcal{B},f}{s}, \pnHierarchicalXXX{j,v,\omega}{\mathcal{B},\send}{s} and \pnHierarchicalXXX{j,v,\omega}{\mathcal{B},\receive}{s}. Each PN's structure depends on the organisation of \transitionFunctionX{j,v,\omega} computations and communication, with pages activated: sequentially, in parallel or in a hybrid manner (defined by the user, potentially combining sequential and parallel structures). Typically PNs: \pnHierarchicalXXX{j,v,\omega}{\mathcal{B},f}{s}, \pnHierarchicalXXX{j,v,\omega}{\mathcal{B},\send}{s} and
\pnHierarchicalXXX{j,v,\omega}{\mathcal{B},\receive}{s} adopt parallel arrangement since operation of pages can be independent, i.e.\ \subsystemX{j,v} can independently send data from each output buffer. However, sending data sequentially can be preferable.
For parallel arrangements, the analysis follows similar rules as for $\mathcal{H}_{j}$
and for sequential arrangements, as the analysis of $^s\mathcal{H}_{j,v,\omega}^{\mathcal{B}}$.
For hybrid configuration, defined by the developer, we follow rules similar to those used for $^s\mathcal{H}_{j,v}$.

%%%%%%%%%%%%%%%%%%%%%%%%%%%%%%%%%%%%%%%%%%%%%%%%%%%%%%%%%%%%%%%%%%%%%%%%%%%%%%%%%%%%%%%%%%%%%%%%%%%%%%%%%%%%%%%%%%%%%%%%
%%%%%%%%%%%%%%%%%%%%%%%%%%%%%%%%%%%%%%%%%%%%%%%%%%%%%%%%%%%%%%%%%%%%%%%%%%%%%%%%%%%%%%%%%%%%%%%%%%%%%%%%%%%%%%%%%%%%%%%%
%%%%%%%%%%%%%%%%%%%%%%%%%%%%%%%%%%%%%%%%%%%%%%%%%%%%%%%%%%%%%%%%%%%%%%%%%%%%%%%%%%%%%%%%%%%%%%%%%%%%%%%%%%%%%%%%%%%%%%%%

\subsection{Analysis of the general communication model}
\label{subsec:rshpn_analysis_general_communication_model}

\begin{figure}
	\centering
	\includegraphics[width=0.95\linewidth]{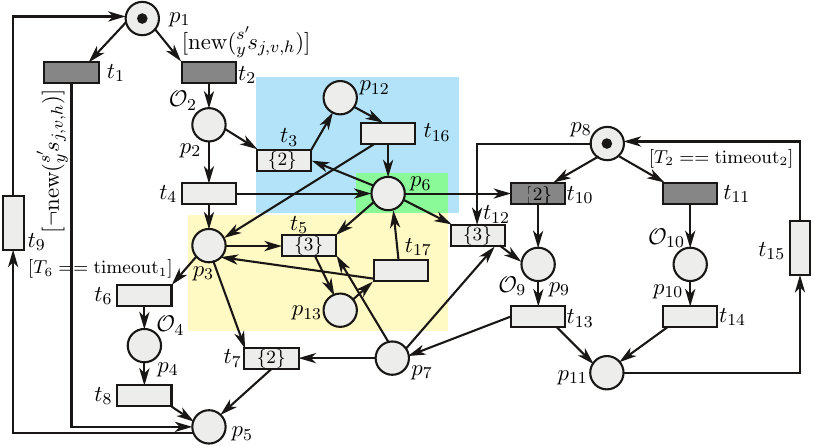}
	\caption{Pure PN (communication model from Fig.~\ref{fig:general_communication_modes_petri_net})}
	\label{fig:general-communication-model-analysis_without_selfloops}
\end{figure}

PNs forming send and receive mode sublayers can not be analysed independently, as
the last sublayer models the interaction between a~pair of communicating subsystems where the PNs are connected by fusion places~\cite{Figat:2022:RAS}.
Both nets form a~single PN representing the communication model for a~pair of subsystems.
Sec.~\ref{subsec:rshpn-communication-models} introduces 9 different communication models, however, all of them can be represented by a~single PN (Fig.~\ref{fig:general_communication_modes_petri_net}).
Thus a~single PN is analysed.
To produce its incidence matrix from the network in Fig.~\ref{fig:general_communication_modes_petri_net} three self-loops must be removed:
1) $t_3 \leftrightarrow p^{\rm fusion}_{(1,1)}$,
2) $t_8 \leftrightarrow p_3$, and
3) $t_8 \leftrightarrow p^{\rm fusion}_{(1,1)}$.
The result is a~pure PN, i.e.\ without self-loops, (Fig.~\ref{fig:general-communication-model-analysis_without_selfloops}).
This network is transformed into incidence matrix $N^{\rm model}$ (Eq.~(\ref{eq:analysis-petri-net-matrix-general-communication-model})) without  losing information about the network operation.
\begin{equation}
	\renewcommand{\arraystretch}{0.7}
	{\scriptsize %\footnotesize
			\begin{blockarray}{*{12}{c@{\hspace{8pt}}}cc}
				\multicolumn{3}{c}{{\footnotesize $N^{\rm model}=$}} \\
				\pnPlaceX{1} & \pnPlaceX{2} & \pnPlaceX{3} & \pnPlaceX{4} & \pnPlaceX{5} & \pnPlaceX{6} & \pnPlaceX{7} & \pnPlaceX{8} & \pnPlaceX{9} & \pnPlaceX{10} & \pnPlaceX{11} & \pnPlaceX{12} & \pnPlaceX{13} \\
				\begin{block}{(*{12}{c@{}}c)c}
					%	1   2   3   4   5   6   7   8   9  10  11  12  13
					-1& 0 & 0 & 0 & 1 & 0 & 0 & 0 & 0 & 0 & 0 & 0 & 0  &\pnTransitionX{1}\\
					-1& 1 & 0 & 0 & 0 & 0 & 0 & 0 & 0 & 0 & 0 & 0 & 0  &\pnTransitionX{2}\\
					0 &-1 & 0 & 0 & 0 &-1 & 0 & 0 & 0 & 0 & 0 & 1 & 0  &\pnTransitionX{3}\\
					%	1   2   3   4   5   6   7   8   9  10  11  12  13
					0 &-1 & 1 & 0 & 0 & 1 & 0 & 0 & 0 & 0 & 0 & 0 & 0  &\pnTransitionX{4}\\
					0 & 0 &-1 & 0 & 0 &-1 &-1 & 0 & 0 & 0 & 0 & 0 & 1  &\pnTransitionX{5}\\
					0 & 0 &-1 & 1 & 0 & 0 & 0 & 0 & 0 & 0 & 0 & 0 & 0  &\pnTransitionX{6}\\
					%	1   2   3   4   5   6   7   8   9  10  11  12  13
					0 & 0 &-1 & 0 & 1 & 0 &-1 & 0 & 0 & 0 & 0 & 0 & 0  &\pnTransitionX{7}\\
					0 & 0 & 0 &-1 & 1 & 0 & 0 & 0 & 0 & 0 & 0 & 0 & 0  &\pnTransitionX{8}\\
					1 & 0 & 0 & 0 &-1 & 0 & 0 & 0 & 0 & 0 & 0 & 0 & 0  &\pnTransitionX{9}\\
					%	1   2   3   4   5   6   7   8   9  10  11  12  13
					0 & 0 & 0 & 0 & 0 &-1 & 0 &-1 & 1 & 0 & 0 & 0 & 0  &\pnTransitionX{10}\\	
					0 & 0 & 0 & 0 & 0 & 0 & 0 &-1 & 0 & 1 & 0 & 0 & 0  &\pnTransitionX{11}\\		
					0 & 0 & 0 & 0 & 0 &-1 &-1 &-1 & 1 & 0 & 0 & 0 & 0  &\pnTransitionX{12}\\		
					%	1   2   3   4   5   6   7   8   9  10  11  12  13
					0 & 0 & 0 & 0 & 0 & 0 & 1 & 0 &-1 & 0 & 1 & 0 & 0  &\pnTransitionX{13}\\		
					0 & 0 & 0 & 0 & 0 & 0 & 0 & 0 & 0 &-1 & 1 & 0 & 0  &\pnTransitionX{14}\\		
					0 & 0 & 0 & 0 & 0 & 0 & 0 & 1 & 0 & 0 &-1 & 0 & 0  &\pnTransitionX{15}\\			
					%	1   2   3   4   5   6   7   8   9  10  11  12  13
					0 & 0 & 1 & 0 & 0 & 1 & 0 & 0 & 0 & 0 & 0 &-1 & 0  &\pnTransitionX{16}\\		
					0 & 0 & 1 & 0 & 0 & 1 & 0 & 0 & 0 & 0 & 0 & 0 &-1  &\pnTransitionX{17}\\
					%	1   2   3   4   5   6   7   8   9  10  11  12  13
				\end{block}
			\end{blockarray}
		}
		\label{eq:analysis-petri-net-matrix-general-communication-model}
	\end{equation}
	Eq.~(\ref{eq:transition_invariant_equation_solution_general_communication_model}) (transition invariant)
	and Eq.~(\ref{eq:place_invariant_equation_solution_general_communication_model}) (place invariant) are solved assuming:
	${\bf x}^{\rm model} \neq {\bf 0}$, ${\bf y}^{\rm model} \neq {\bf 0}$ and  restricting the solution to non-negative integers.
	\begin{equation}
		(N^{\rm model})^\top \cdot {\bf x}^{\rm model} = (0, 0, 0, 0, 0, 0, 0, 0, 0, 0, 0, 0, 0)^\top
		\label{eq:transition_invariant_equation_solution_general_communication_model}
	\end{equation}
	\begin{equation}
		N^{\rm model} \cdot {\bf y}^{\rm model} = (0, 0, 0, 0, 0, 0, 0, 0, 0, 0, 0, 0, 0, 0, 0, 0, 0)^\top
		\label{eq:place_invariant_equation_solution_general_communication_model}
	\end{equation}
	%%%%%%%%%%%%%%%%%%%%%%%%%%%%%%%%%%%%%%%%%%%%%%%%%%%%%%%%%%%%
For $x_{\delta}\in\{x_9, x_{12}, x_{14}, x_{15}, x_{16}, x_{17}\}$,	assuming that $x_{\delta}\in\mathbb{Z}$ and $x_{\delta}>0$,  Eq.~(\ref{eq:transition_invariant_equation_solution_communication_model}) is produced, thus determining the transition invariants.
	\begin{minipage}{0.33\textwidth}
	\begin{equation}
		{\footnotesize {\bf x}^{\rm model} =
			\begin{blockarray}{c}
				\begin{block}{(c)}
					x_{9} + x_{14} - x_{15} - x_{16} \\
					-x_{14} + x_{15} + x_{16} \\
					x_{16} \\
					-x_{14} + x_{15} \\
					x_{17} \\				
					x_{12} + x_{16} + x_{17} \\
					-x_{12} - x_{14} + x_{15} - x_{17} \\
					x_{12} + x_{16} + x_{17} \\
					x_{9} \\
					-x_{12} - x_{14} + x_{15} \\
					x_{14} \\
					x_{12} \\
					-x_{14} + x_{15} \\
					x_{14} \\
					x_{15} \\
					x_{16} \\
					x_{17} \\					
				\end{block}
			\end{blockarray}
		} % end of scriptsize
		\label{eq:transition_invariant_equation_solution_communication_model}
	\end{equation}
	\end{minipage}%
	\begin{minipage}{0.16\textwidth}
		\begin{equation}
			{\footnotesize {\bf y}^{\rm model} =
				\begin{blockarray}{c}
					\begin{block}{(c)}
						y_{13} \\
						y_{13} \\
						y_{13} \\
						y_{13} \\
						y_{13} \\					
						0 \\
						0 \\
						y_{11} \\
						y_{11} \\
						y_{11} \\
						y_{11} \\
						y_{13} \\
						y_{13} \\
					\end{block}
				\end{blockarray}
			} % end of scriptsize
			\label{eq:place_invariant_equation_solution_general_communication_model_2}
		\end{equation}
	\end{minipage}
	Now  the positive transition invariant is determined, for which $\left\Vert {\bf x}^{\rm model} \right\Vert$ contains the maximum number of different transitions.
	Setting $x_{9}=5$, $x_{12}=1$, $x_{14}=1$, $x_{15}=4$, $x_{16}=1$, $x_{17}=1$,
	then ${\bf x}^{\rm model} = (1 , 4 , 1 , 3 , 1 , 3 , 1 , 3 , 5 , 2 , 1 , 1 , 3 , 1 , 4 , 1 , 1)^\top$. Thus firing all the transitions as many times as ${\bf x}^{\rm model}$ indicates, the network will return to $M_0$.
	As shown in Sec.~\ref{subsubsec:place_transition_invariant_analysis}, if it is possible to reach $M_0$ by firing all transitions in the PN then the net is live. And since it is live there are no deadlocks.
	Interestingly, by manipulating the parameters $x_{9}$, $x_{12}$, $x_{14}$, $x_{15}$, $x_{16}$ and $x_{17}$ one can obtain different combinations of sender and receiver modes, e.g.:
	\begin{itemize}
		\item if $x_9=1$ (all others equal zero) then $ {\bf x}^{\rm model} = ( 1 , 0 , 0 , 0 , 0 , 0 , 0 , 0 , 1 , 0 , 0 , 0 , 0 , 0 , 0 , 0 , 0 )$. The vector obtained indicates that the sending subsystem does not compute new data and therefore does not send any data. After firing $t_1$ and $t_9$, the network marking returns to the initial $M_0$ marking, i.e. $M_0 \xrightarrow{t_1} \ldots \xrightarrow{t_9} M_0$,
		\item if $x_9=1$, $x_{15}=1$, $x_{17}=1$ then ${\bf x}^{\rm model} = ( 0 , 1 , 0 , 1 , 1 , 0 , 1 , 0 , 1 , 1 , 0 , 0 , 1 , 0 , 1 , 0 , 1 )$.	If the transitions indicated in the vector fire , e.g. in the order: $t_2 \rightarrow t_4\rightarrow t_{10} \rightarrow t_{13} \rightarrow t_{5} \rightarrow t_{17} \rightarrow t_7\rightarrow t_9 \rightarrow t_{15}$, then successful communication will take place between the sender and receiver (both operating in blocking mode),		
		\item if $x_9=2$,  $x_{14}=1$,  $x_{15}=2$, $x_{16}=1$, $x_{17}=1$ then ${\bf x}^{\rm model} = ( 0 , 2 , 1 , 1 , 1 , 2 , 0 , 2 , 2 , 1 , 1 , 0 ,$ $1 , 1 , 2 , 1 , 1 )$. If the transitions indicated in the vector fire, e.g. in the order:
		$t_2 \rightarrow t_4 \rightarrow t_6 \rightarrow t_8 \rightarrow t_9 \rightarrow t_2 \rightarrow t_3 \rightarrow t_{16} \rightarrow t_{10} \rightarrow t_{13} \rightarrow t_{15} \rightarrow t_{5} \rightarrow t_{17} \rightarrow t_{6} \rightarrow t_{8} \rightarrow t_{9} \rightarrow t_{11} \rightarrow t_{14} \rightarrow t_{15}$ then communication will take place between the two subsystems operating in non-blocking mode.
	\end{itemize}

	%%%%%%%%%%%%%%%%%%%%%%%%%%%%%%%%%%
	Eq.~(\ref{eq:place_invariant_equation_solution_general_communication_model_2}), where: $y_{11},y_{13}\in\mathbb{Z}$ and $y_{11},y_{13}>0$,  represents the solution of Eq.~(\ref{eq:place_invariant_equation_solution_general_communication_model}). The vector ${\bf y}^{\rm model}$ determines all possible place invariants. Two linearly independent place invariants can be derived:
	a) $^1{\bf y}^{\rm model} = (1, 1, 1, 1, 1, 0, 0, 0, 0, 0, 0, 1, 1)^\top$ for $y_{13}=1 \land y_{11}=0$, and
	b) $^2{\bf y}^{\rm model} = (0, 0, 0, 0, 0, 0, 0, 1, 1, 1, 1, 0, 0)^\top$ for $y_{13}=0 \land  y_{11}=1$, thus the weighted sum of the tokens remains constant for each reachable marking $M_m$, i.e: 				$M_{m}^\top \cdot\ ^1{\bf y}^{\rm model} = M_{0}^\top \cdot\ ^1{\bf y}^{\rm model} = 1$ and $M_{m}^\top \cdot\ ^2{\bf y}^{\rm model} = M_{0}^\top \cdot\ ^2{\bf y}^{\rm model} = 1$, where $M_0=(1, 0, 0, 0, 0, 0, 0, 1, 0, 0, 0, 0, 0)^\top$.
	From the above two equations, Eq.~(\ref{eq:place_invariant_equation_solution_general_communication_model_2_4a}) and Eq.~(\ref{eq:place_invariant_equation_solution_general_communication_model_2_4b}) are derived, which are satisfied for any reachable $M_m$.
	\begin{equation}
		M_m(p_1) + \ldots + M_m(p_5) + M_m(p_{12}) +  M_m(p_{13})= 1
		\label{eq:place_invariant_equation_solution_general_communication_model_2_4a}
	\end{equation}
	\begin{equation}
		M_m(p_8) + M_m(p_9) + M_m(p_{10}) + M_m(p_{11}) = 1
		\label{eq:place_invariant_equation_solution_general_communication_model_2_4b}
	\end{equation}

	It follows from Eq.~(\ref{eq:place_invariant_equation_solution_general_communication_model_2_4a}) that the sum of the tokens in $p_1$, $p_2$, $p_3$, $p_4$, $p_5$, $p_{12}$ and $p_{13}$ always equals 1, similarly, for $p_8$, $p_9$, $p_{10}$, $p_{11}$ based on Eq.~(\ref{eq:place_invariant_equation_solution_general_communication_model_2_4b}).
	Then, to prove that the PN (Fig.~\ref{fig:general-communication-model-analysis_without_selfloops}) is safe, it suffices to show that for any marking $M_m$:
	$M_m(\pnPlaceX{6})\le1$, and
	$M_m(\pnPlaceX{7})\le1$ (since the condition for the remaining places is satisfied, as follows from  Eq.~(\ref{eq:place_invariant_equation_solution_general_communication_model_2_4a}) and Eq.~(\ref{eq:place_invariant_equation_solution_general_communication_model_2_4b})).
	To prove the first one, a~single token inserted into \pnPlaceX{6} by either \pnTransitionX{16} (requires prior firing of $t_{3}$) or \pnTransitionX{4}.
	The choice which transition fires, when $M_m(\pnPlaceX{2})=1$, depends on the current marking of \pnPlaceX{6}.
	If $M_m(\pnPlaceX{6})=0$ then \pnTransitionX{4} fires producing a~single token in \pnPlaceX{6}, i.e.\ as a~result $M_m(\pnPlaceX{6})=1$.
	Otherwise, if $M_m(\pnPlaceX{6})=1$ then \pnTransitionX{3} fires consuming a~token from \pnPlaceX{6} and produces a~token in \pnPlaceX{12}. Subsequently, $t_{16}$ fires producing a~token in $p_{6}$. As a~result the number of tokens in $p_6$ remains the same.
	Thus, tokens will never multiply in \pnPlaceX{6} and $M_m(\pnPlaceX{6})\le1$.
	Similar inference is made for  \pnPlaceX{7} based on \pnTransitionX{12} and \pnTransitionX{13}. Transitions \pnTransitionX{3} and \pnTransitionX{12} assure that the tokens do not multiply within \pnPlaceX{6} and \pnPlaceX{7}, respectively (if the token was not consumed by \pnTransitionX{10} and \pnTransitionX{7}). Summarizing, for each \pnPlaceX{i} ($i=1,\dots,13$) and any marking $M_m$, the $M_m(\pnPlaceX{i})\le1$ is always fulfilled, thus the PN is safe.
	
	The obtained place invariants are used to check whether the PN is conservative. It is sufficient to sum the two obtained minimal place invariants, i.e.\ $^1{\bf y}^{\rm model}$ and $^2{\bf y}^{\rm model}$. This yields a  vector of non-negative weights $(1,1,1,1,1,0,0,1,1,1,1,1,1)^\top$, for which the weighted number of tokens in any reachable marking is equal to 2. Thus the PN is partially conservative. Moreover, it can be demonstrated that the PN can perform two operations simultaneously, i.e.\ sending from $_{y}s_{j,v}$ and receiving into $_{x}s_{j',h}$.
	
	%%%%%%%%%%%%%%%%%%%%%%%%%%%%%%%%%%%%%%%%%%%%%%%%%%%%%%%%%%%%%%%%%%%%%%%%%%%%%%%%
	
	%%%%%%%%%%%%%%%%%%%%%%%%	
	
	 % OK
% ====  END - OK ======

% !TeX root = ../figat_analiza.tex

\section{Observations Regarding Analysis Methods}
\label{sec:results}

\subsection{Properties preservation}
\label{subsec:properties-preservation}
	%{\bf Properties preservation:}
	RSHPN meta-model analysis confirms \pnHierarchicalX{} is safe and conservative. Table~\ref{tab:petri_net_analysis_summary_conservativeness} summarizes meta-model properties preservation for each layer. The first column names the layer; the second indicates if the net is user-defined ('yes') or derived from the meta-model ('no') -- where 'yes' requires full PN analysis and 'no' preserves meta-model properties. The third column shows conservativeness preservation, the fourth lists the vector of weights $\bf y$, and the fifth gives the weighted sum of tokens. In \pnHierarchicalX{}, subnets include: (1)~strictly conservative -- performing one operation at a time, (2)~conservative -- performing multiple operations simultaneously, or (3)~partially conservative -- performing exactly two operations. If a~PN fails to produce the correct weighted sum, it is incompatible with the meta-model and must be corrected. This enables the automation of RSHPN analysis. Table~\ref{tab:petri_net_analysis_summary_conservativeness} indicates that RSHPN model analysis focuses on networks with a task-dependent (variable) structure (layers with 'yes'): 1) produced by the designer, i.e.\ those in subsystem layer or having hybrid arrangement, and 	2) modelling the interaction between communicating subsystems (only if B-B mode is used). This property of RSHPN significantly reduces the analysis effort. If all subsystems of \roboticSystem\ communicate in non-blocking mode, and all operations performed by the subsystems are either sequential or parallel, the analysis of the developed \pnHierarchicalX{} model further reduces to the analysis of the networks in the subsystem layer only. Properties of PNs in other layers are preserved from the meta-model since these PNs have the same structure.

\begin{table}
\centering
\caption{Summary of RSHPN \pnHierarchicalX{} analysis regarding conservativeness (where $\zeta = |{_{y}\hat{s}_{j,v}}|$, $\nu = |{_{x}\hat{s}_{j,v}}|$)}
\scalebox{0.85}{	
	\renewcommand{\arraystretch}{1.35}
	\begin{tabular}{|c|c|c|c|c|}
		\hline
		\multirow{2}{1.3cm}{PN} & \multirow{2}{0.7cm}{User defined} & \multirow{2}{{1.5cm}}{Conservative} & \multirow{2}{1.25cm}{Vector $\bf y$} & \multirow{2}{0.9cm}{Weighted sum \mfb{$k$}}\\
		\ & \ & \ & \ & \ \\ % Zastąpione puste miejsca przez '\ ' dla zachowania krawędzi
		
		\hline \hline
		\pnHierarchicalX{} & no & yes & $(\setSize{\agentSet{}},1,\ldots,1)$ & \setSize{\agentSet{}} \\ \hline
		\pnHierarchicalX{j} & no & yes & $(\setSize{\subsystemSetX{j}},1,\ldots,1,\setSize{\subsystemSetX{j}})$ & \setSize{\subsystemSetX{j}}\\ \hline
		\pnHierarchicalXXX{j,v}{}{s} & yes & strictly & $(1,\ldots,1)$ & 1\\ \hline
		\pnHierarchicalXXX{j,v,\omega}{\mathcal{B}}{s} & no & strictly & $(1,\ldots,1)$ & 1 \\ \hline

		\multicolumn{3}{c}{\pnHierarchicalXXX{j,v,\omega}{\mathcal{B},f}{s}, \pnHierarchicalXXX{j,v,\omega}{\mathcal{B},\rm send}{s},
			\pnHierarchicalXXX{j,v,\omega}{\mathcal{B},\rm rcv}{s}:} \\ \hline
		sequential & no & strictly & $(1,\ldots,1)$ & 1 \\ \hline
		\multirow{1}{*}{parallel} & \multirow{1}{*}{no} & \multirow{1}{*}{yes} & \multirow{1}{*}{$(\zeta,1,\ldots,1,\zeta)$} & $\nu$ \\ \hline
		hybrid & yes & yes & \multicolumn{2}{c|}{depends on the structure}\\ \hline \hline
		\pnHierarchicalXXX{j,v,\omega, \psi}{\mathcal{B},f}{s} & no & strictly & $(1,\ldots,1)$ & 1 \\ \hline
		\pnHierarchicalXXX{j,v,\omega, \rho}{\mathcal{B},\rm snd}{s} & \multirow{2}{*}{no} & \multirow{2}{*}{partially} & $(1,1,1,1,1,$ &\multirow{2}{*}{2} \\
		\pnHierarchicalXXX{j,v,\omega,\kappa}{\mathcal{B},\rm rcv}{s}& & &$0,0,1,1,1,1)$& \\ \hline	
	\end{tabular}
}
\label{tab:petri_net_analysis_summary_conservativeness}
\end{table}

\subsection{State space explosion}
\label{subsec:state-space-explosion}
%{\bf State space explosion:}
Assuming each agent has the same number of subsystems and each subsystem has the same number of behaviours, input and output buffers, % Eq.~(\ref{eq:place_robot_whole})
the number of places in $\mathcal{H}$ is approximated by:
\begin{align}
	\tiny
	\placesInHPNX{}^{\rm simple} = \setSize{\agentSet} (\setSize{\subsystemSet_{j}} \cdot \setSize{\behaviourSetX{\agentIndex, \subsystemIndex}} \cdot
	(2^{\setSize{\inputBufferSetX{\agentIndex, \subsystemIndex}}} \cdot (\setSize{\outputBufferSetX{\agentIndex, \subsystemIndex}}+1) + \nonumber \\
	+ 7\cdot\setSize{\inputBufferSetX{\agentIndex, \subsystemIndex}}
	+ 11\cdot\setSize{\outputBufferSetX{\agentIndex, \subsystemIndex}} + 16)
	+ 3\cdot\setSize{\subsystemSetX{\agentIndex}})
	+ 3\cdot\setSize{\agentSet} + 1
	\label{eq:place_robot_whole}
\end{align}
where: $\agentSet$, $\subsystemSet{_j}$, $\behaviourSetX{\agentIndex, \subsystemIndex}$, $\inputBufferSetX{\agentIndex, \subsystemIndex}$, $\outputBufferSetX{\agentIndex, \subsystemIndex}$ are sets of: agents, subsystems in each agent $a_j$, behaviours in $s_{j,v}$ and input and output buffers, respectively. For $\mathcal{RS}$ with $\setSize{\subsystemSet_{j}}$ subsystems operating in parallel for each $a_{j} \in \agentSet$, the estimated number of states in the reachability graph depends on the exponent ${\setSize{\subsystemSet_j}\cdot \setSize{\agentSet}}$: 
\begin{align}
	\tiny
	{\rm states\_in\_reachability\_graph}(\mathcal{H})	= [
	\setSize{\behaviourSetX{\agentIndex, \subsystemIndex}} \cdot (2^{\setSize{\inputBufferSetX{\agentIndex, \subsystemIndex}}} \nonumber \\ \cdot (\setSize{\outputBufferSetX{\agentIndex, \subsystemIndex}} +1)
	+ 7\cdot\setSize{\inputBufferSetX{\agentIndex, \subsystemIndex}}+ 11\cdot\setSize{\outputBufferSetX{\agentIndex, \subsystemIndex}}+16) + 3\cdot\setSize{\subsystemSetX{\agentIndex}}
	] ^{\setSize{\subsystemSet_j}\cdot \setSize{\agentSet}}
	\label{eq:states_in_reachability_graph}
\end{align}
This leads to an exponential increase in the number of possible states in the reachability graph, making it infeasible to analyse even for relatively simple systems using standard methods. An example of table tennis balls collecting robot illustrates this~\cite{Figat:2022:RAS}. In this article, we propose decomposing RSHPN analysis into the analysis of individual PNs in panels. As the number of panels depends on agents, subsystems, and behaviours (e.g., the estimated number of panels in the first four layers is $1 + \setSize{\hat{a}} + \setSize{\hat{a}} \cdot \setSize{\hat{s}_{j}} + \setSize{\hat{a}} \cdot \setSize{\hat{s}_{j}} \cdot \setSize{\hat{\mathcal{B}}_{j,v}}$), the number of necessary states in the reachability graphs no longer depends on the exponent ${\setSize{\subsystemSet_j} \cdot \setSize{\agentSet}}$. This key contribution ensures only user-dependent layers, such as the subsystem layer and those with hybrid arrangements, need the analysis, while others preserve the meta-model properties.

\subsection{\mfb{PN complexity assessment}}
\label{subsec:pn-complexity-assessment}

\mfb{The complexity of an RSHPN network can be assessed in three ways: (1) structural complexity, (2) token variability, and (3) the relationship between structure and state-space complexity.}

\mfb{Structural complexity is expressed as the sum of places, transitions, and edges in the network (Eq.~\ref{eq:model-complexity_final2_analysis}).
\begin{equation}
	\resizebox{1.0\hsize}{!}{$
		\begin{array}{l}
			{O}^{\rm model}({\rm RSHPN}) =
			\sum_{ \agentj{\agentIndex} \in \agentSet} \big[\sum_{ \sbbbj{\agentIndex,\subsystemIndex} \in \subsystemSetX{\agentIndex}} \big[\setSize{\behaviourSetX{\agentIndex, \subsystemIndex}} \cdot \big(  112\cdot\setSize{\outputBufferSetX{\agentIndex, \subsystemIndex}}
			+ 81\cdot \setSize{\inputBufferSetX{\agentIndex, \subsystemIndex}} +\\
			+ 2^{\setSize{\inputBufferSetX{\agentIndex, \subsystemIndex}}} \cdot \big( 21\cdot\setSize{\outputBufferSetX{\agentIndex, \subsystemIndex}} +
			23\big) + 148\big) +
			10\cdot\setSize{\initialConditionSetX{\agentIndex, \subsystemIndex}}\big]
			+35 \cdot \setSize{\subsystemSetX{\agentIndex}} \big]	+ 29\cdot \setSize{\agentSet} + 14
			\label{eq:model-complexity_final2_analysis}
		\end{array}$
	}
\end{equation}
where $\initialConditionSetX{\agentIndex, \subsystemIndex}$ represents the set of initial conditions~\cite{Figat:2022:RAL}. To reduce computational overhead, we limit overloaded functions to at the most two (thus, $2^{\setSize{\inputBufferSetX{\agentIndex, \subsystemIndex}}} = 2$). However, even simple networks can exhibit high complexity with an infinite number of reachable states, e.g., $p_1 \rightarrow t_1 \xrightarrow{2} p_1$. Consequently, structural complexity analysis alone is not always sufficient.}

\mfb{It is essential to examine the variability of tokens in PN, which is directly related to its conservativeness. Table~\ref{tab:petri_net_analysis_summary_conservativeness} summarizes this assessment for RSHPN layers:
1)~\textbf{Strictly conservative}: the weighted sum of tokens remains constant, as incoming and outgoing edges are balanced;
2)~\textbf{Conservative}: some transitions modify the weighted token sum, but the system can return to the initial marking;
3)~\textbf{Partial/non-conservative}: tokens proliferate, making state-space growth unpredictable. The conservativeness of a network can be further quantified using the weighted sum of tokens, defined as $k = M_0^\top \cdot \mathbf{y}$, where $k$ represents the total weighted token count, and $\mathbf{y}$ defines the weight of each place. The parameter $k$ provides insight into the complexity of token variability: higher values suggest richer network dynamics, where tokens are more freely distributed among places, often leading to more possible system states. This is particularly relevant in networks with extensive branching structures, where tokens can follow multiple independent execution paths. In \textbf{strictly conservative networks} ($\forall_{p_i} {\bf y}(p_i) = 1$), the sum of weighted token values is exactly preserved across all token configurations, leading to a limited state-space: $|\hat{M}| = |\hat{p}|$. For \textbf{general conservative networks} ($\forall_{p_i} {\bf y}(p_i) > 0$), the weighted sum remains constant, though individual token distributions may change. The number of reachable states is then bounded by $|\hat{M}| \leq \binom{|\hat{p}| + k - 1}{k}$, which follows from combinatorial theory, where a total weight $k$ is distributed across $|\hat{p}|$ distinguishable places. The binomial coefficient accounts for all possible redistributions that maintain the conservation constraint. A particularly important case arises when some entries in $\mathbf{y}$ are zero ($\exists p_i: {\bf y}(p_i) = 0$). In such cases, the number of tokens in these places does not affect the total weighted sum, meaning that tokens can accumulate without bound. This creates the potential for uncontrolled token growth, significantly increasing the variability of network states. These places act as unregulated expansion points, where tokens can proliferate freely, leading to state-space explosion.}

\mfb{In summary, reachable states can be accurately determined, but only with full knowledge of the network structure. While combinatorial approximations provide useful upper bounds, they do not capture the full complexity of token interactions in non-trivial networks. Structural reductions and equation-based methods~\cite{Berthomieu2020} provide accurate estimates, but require detailed network analysis, making them impractical for large systems. Thus, both $k$ and the structure of $\mathbf{y}$ are crucial in determining the complexity of token variability in a network. A higher $k$ suggests that tokens are more widely redistributed, increasing network complexity, while places with ${\bf y}(p_i) = 0$ indicate the possibility of unbounded token growth, resulting in an unpredictable and rapidly expanding state space.}

%%%%%%%%%%%%%%%%%%%%

%%%%%%%%%%%%%%%%%%%%%%%%%%%%%%%%%%%%%%%%%%%%%%%%%%%%%%%%%%%%%%%%%%%%%%%%%%%%%%%%%%%%%%%%%%%%%%%%%%%%%%%%%%%%%%%%%%%%%%%%%%%%%%%%%%%%%%%%%%%%%%%%%%%%%%%%%%%%%%%%
%%%%%%%%%%%%%%%%%%%%%%%%%%%%%%%%%%%%%%%%%%%%%%%%%%%%%%%%%%%%%%%%%%%%%%%%%%%%%%%%%%%%%%%%%%%%%%%%%%%%%%%%%%%%%%%%%%%%%%%%%%%%%%%%%%%%%%%%%%%%%%%%%%%%%%%%%%%%%%%%
%%%%%%%%%%%%%%%%%%%%%%%%%%%%%%%%%%%%%%%%%%%%%%%%%%%%%%%%%%%%%%%%%%%%%%%%%%%%%%%%%%%%%%%%%%%%%%%%%%%%%%%%%%%%%%%%%%%%%%%%%%%%%%%%%%%%%%%%%%%%%%%%%%%%%%%%%%%%%%%%
%%%%%%%%%%%%%%%%%%%%%%%%%%%%%%%%%%%%%%%%%%%%%%%%%%%%%%%%%%%%%%%%%%%%%%%%%%%%%%%%%%%%%%%%%%%%%%%%%%%%%%%%%%%%%%%%%%%%%%%%%%%%%%%%%%%%%%%%%%%%%%%%%%%%%%%%%%%%%%%%
%%%%%%%%%%%%%%%%%%%%%%%%%%%%%%%%%%%%%%%%%%%%%%%%%%%%%%%%%%%%%%%%%%%%%%%%%%%%%%%%%%%%%%%%%%%%%%%%%%%%%%%%%%%%%%%%%%%%%%%%%%%%%%%%%%%%%%%%%%%%%%%%%%%%%%%%%%%%%%%%

\subsection{\mfb{Decision on PN analysis method selection}}
\label{subsec:decision-on-pn-analysis-method-selection}

\mfb{Once the structural complexity and token variability of a PN have been assessed, the next step is to select an appropriate analysis method. A key factor in this decision is the estimated number of reachable states relative to the network’s structural complexity: $\frac{|\hat{p}| + |\hat{t}| + |\hat{e}|}{|\hat{M}|}$, where $\hat{e}$ is the set of edges and $\hat{M}$ is the set of reachable states. If this ratio is very low (meaning the number of reachable states grows exponentially compared to network size), the invariant-based method is the most suitable choice. Conversely, if the number of reachable states remains manageable, the reachability graph method or graphical interpretation method can be applied. Table~\ref{tab:analysis_methods} presents a comparison of the three primary PN analysis methods.}

\begin{table}[h]
	\centering
	\caption{Comparison of PN Analysis Methods}
	\scalebox{0.85}{
		\begin{tabular}{|p{1.25cm}|p{1.25cm}|p{1.25cm}|p{2.5cm}|p{1.75cm}|}
			\hline
			\textbf{Method} & \textbf{Considers initial marking?} & \textbf{Considers full state space?} & \textbf{Main application} & \textbf{Drawbacks} \\
			\hline
			Invariants & No & No & Structural analysis, system behaviour independent of initial marking & Does not detect state explosion \\
			\hline
			Reachability graph & Yes & Yes & Full analysis of reachable markings, deadlock detection & State explosion problem \\
			\hline
			PN graphical interpretation & Yes & Partially & Intuitive analysis of problematic places, e.g., growth of the number of tokens & Subjective, lacks formal rigor \\
			\hline
		\end{tabular}
	}
	\label{tab:analysis_methods}
\end{table}

\mfb{As shown in Table~\ref{tab:analysis_methods}, the invariants method does not take the initial marking into account, i.e.\ the number of tokens in the network does not influence the analysis results -- only the network structure matters. This is a key difference between the invariants and reachability graph methods. In the latter case, the initial marking fully determines the set of possible states.}

\mfb{Both reachability graph analysis and place/transition invariants analysis have their respective advantages and limitations; however, they are not mutually exclusive but rather complementary. The primary drawback of reachability graphs is their exponential growth, which leads to state explosion. In contrast, invariants analysis avoids constructing full reachability trees, making it scalable for large models. This makes the invariants-based approach particularly useful for analysing complex systems where exhaustive reachability exploration is infeasible.}

\mfb{While reachability graph analysis explores transitions and system behaviour, invariants analysis focuses on structural properties. This enables verification of cyclic activities without requiring full state-space exploration, making it a powerful tool for assessing global system properties. The reachability graph method is well-suited for verifying state-specific properties such as liveness and deadlocks, whereas invariants analysis ensures that system-wide constraints are satisfied.}

\mfb{Place-invariants analysis identifies subsets of places where the weighted sum of tokens remains constant, revealing fundamental properties of the designed multi-agent robotic system. Transition-invariants analysis determines multisets of transitions whose firings return the PN to its initial marking, capturing essential communication and coordination patterns. By combining these approaches, a more comprehensive understanding of the entire system can be obtained, balancing behavioural and structural perspectives.}

%%%%%%%%%%%%%%%%%%%%%%%%%%%%%%%%%%%%%%%%%%%%%%%%%%%%%%%%%%%%%%%%%%%%%%%%%%%%%%%%%%%%%%%%%%%%%%%%%%%%%%%%%%%%%%%%%%%%%%%%%%%%%%%%%%%%%%%%%%%%%%%%%%%%%%%%%%%%%%%%
%%%%%%%%%%%%%%%%%%%%%%%%%%%%%%%%%%%%%%%%%%%%%%%%%%%%%%%%%%%%%%%%%%%%%%%%%%%%%%%%%%%%%%%%%%%%%%%%%%%%%%%%%%%%%%%%%%%%%%%%%%%%%%%%%%%%%%%%%%%%%%%%%%%%%%%%%%%%%%%%
%%%%%%%%%%%%%%%%%%%%%%%%%%%%%%%%%%%%%%%%%%%%%%%%%%%%%%%%%%%%%%%%%%%%%%%%%%%%%%%%%%%%%%%%%%%%%%%%%%%%%%%%%%%%%%%%%%%%%%%%%%%%%%%%%%%%%%%%%%%%%%%%%%%%%%%%%%%%%%%%
%%%%%%%%%%%%%%%%%%%%%%%%%%%%%%%%%%%%%%%%%%%%%%%%%%%%%%%%%%%%%%%%%%%%%%%%%%%%%%%%%%%%%%%%%%%%%%%%%%%%%%%%%%%%%%%%%%%%%%%%%%%%%%%%%%%%%%%%%%%%%%%%%%%%%%%%%%%%%%%%
%%%%%%%%%%%%%%%%%%%%%%%%%%%%%%%%%%%%%%%%%%%%%%%%%%%%%%%%%%%%%%%%%%%%%%%%%%%%%%%%%%%%%%%%%%%%%%%%%%%%%%%%%%%%%%%%%%%%%%%%%%%%%%%%%%%%%%%%%%%%%%%%%%%%%%%%%%%%%%%%

% !TeX root = ../figat_analiza.tex

%\clearpage

\section{Case study: System Analysis}
\label{sec:case-study}

\begin{figure}[tb]
	\begin{subfigure}{0.34\linewidth}
		\includegraphics[width=1.0\linewidth]{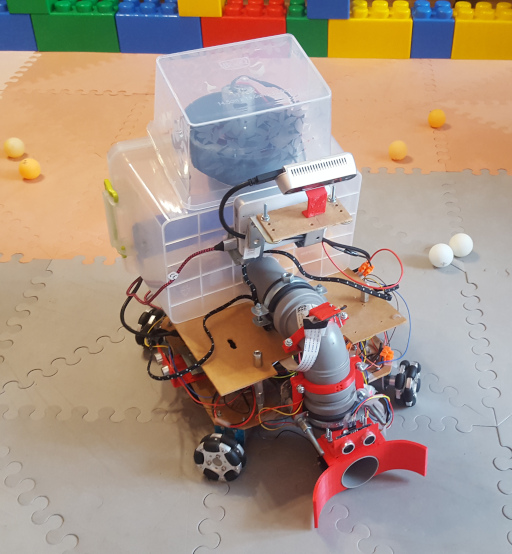}
		\label{fig:robot_collecting_ball}
	\end{subfigure}
	\begin{subfigure}{0.64\linewidth}
		\includegraphics[width=1.0\linewidth]{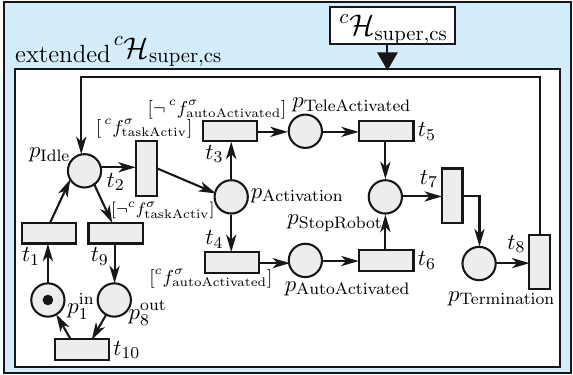}
		\label{fig:supersubsystemanalysis}
	\end{subfigure}
	\centering
	\caption{Left: Robot collecting balls; right: extended \pnHierarchicalXXX{\rm super, cs}{}{c} modelling the activities of \controlSubsystemX{\rm super}.}
	\label{fig:robot_and_super_agent}
\end{figure}

A non-trivial robotic system for collecting table tennis balls~\cite{Figat:2022:RAS} (from Fig.~\ref{fig:robot_and_super_agent}) was developed using the approach based on the RSHPN parametric meta-model. The system has a variable structure.

{\bf Task description:}
The main task of this robotic system is to collect table-tennis balls scattered on a~flat surface. It is required that the robot operates in either of two modes: autonomous or teleoperated. The robot switches between them based on the recognised voice commands uttered by the operator, and terminates its operation when the termination command is issued. In the autonomous mode the robot, using its receptors and effectors, autonomously moves in the environment collecting table-tennis balls, while in the teleoperated mode the robot is controlled by operator commands.

\begin{figure}[b]
	\centering
	\includegraphics[width=0.75\linewidth]{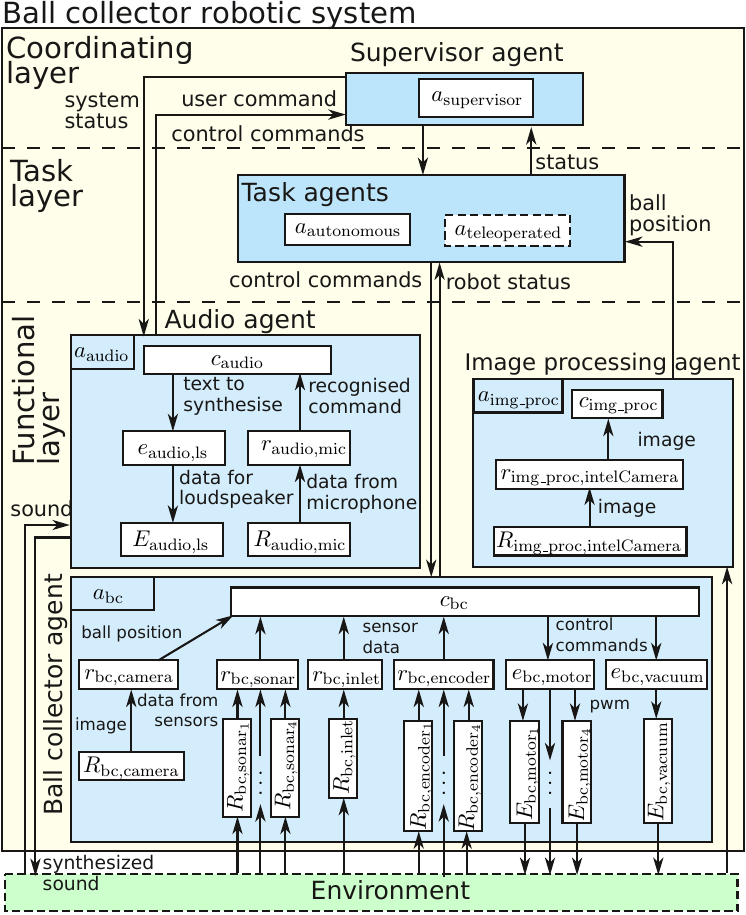}
	\caption{Structure of the ball collector robotic system \mfb{(adapted from~\cite{Figat:2022:RAS})}}
	\label{fig:ballcollectornew4}
\end{figure}

{\bf System structure:}
The structure of the robotic system is presented in Fig.~\ref{fig:ballcollectornew4}. It consists of 6 agents, out of which 4 are always active:
1)~\agentX{\rm audio} -- responsible for sound synthesis and user voice command recognition,
2)~\agentX{\rm super} -- supervisor agent responsible for task agent management and for transferring of recognised control commands,
3)~\agentX{\rm img\_proc} -- responsible for table-tennis ball detection based on neural networks, and
4)~\agentX{\rm bc} -- controlling the robot hardware based on commands received from the task agents. Two other agents: \agentX{\rm auto} (autonomous agent) and \agentX{\rm tele} (teleoperated agent), are activated by \agentX{\rm super} when the user issues voice-commands. The task agents are responsible for task execution. Only one of them can be active at a time.

\begin{figure}
	\centering
	\includegraphics[width=1.0\linewidth]{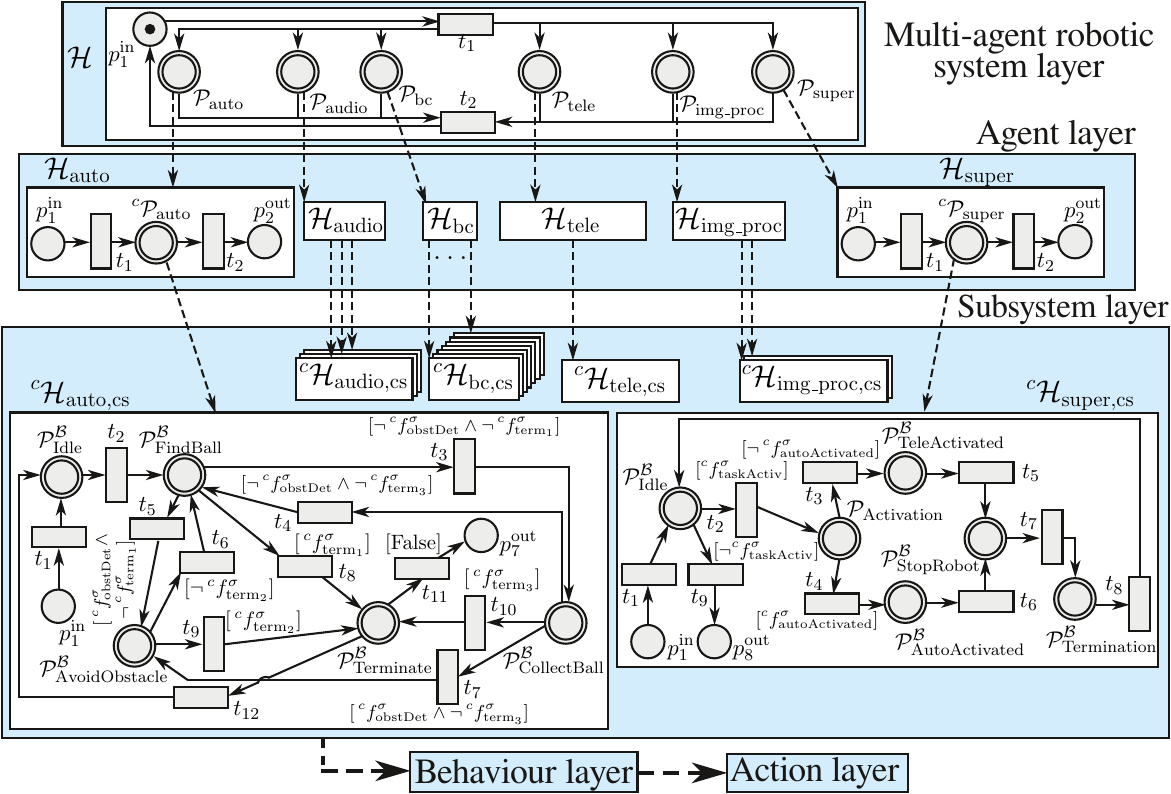}
	\caption{RSHPN \pnHierarchicalXXX{}{}{} modelling the activity of the robotic system (Fig.~\ref{fig:ballcollectornew4}) \mfb{ -- adapted from~\cite{Figat:2022:RAS}}}
	\label{fig:ballcollectorrobotrshpnnew2}
\end{figure}

{\bf RSHPN model:}
The activity of the whole robotic system, modelled using a single RSHPN \pnHierarchicalXXX{}{}{}, is presented in Fig.~\ref{fig:ballcollectorrobotrshpnnew2}. The RSHPN has 1168 places and 1294 transitions, resulting in over 25 million reachable markings (states). Our previous approach to analysis, i.e. which was to flatten the PN and use state-of-the-art analysis tools (such as the Tina tool), proved to be inadequate. The results presented in this article show that for RSHPN it is possible to significantly reduce the complexity of the  analysis.

\begin{figure}[tbh]
	\centering
	\includegraphics[width=0.85\linewidth]{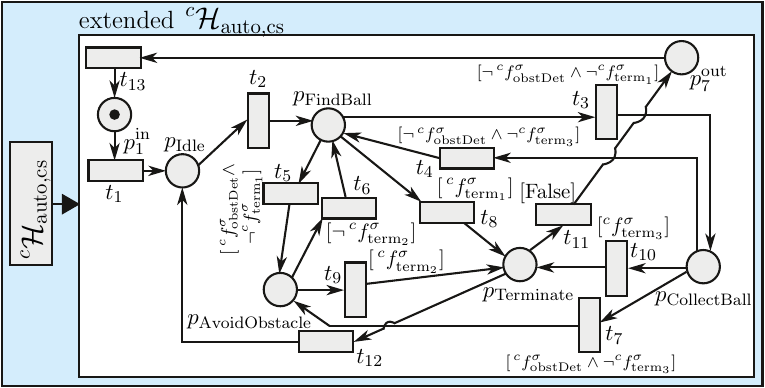}
	\caption{Extended \pnHierarchicalXXX{\rm auto, cs}{}{c} modelling the activities of \controlSubsystemX{\rm auto}.}
	\label{fig:autonomoussubsystemanalysis}
\end{figure}

{\bf Analysis:} As shown in Table~\ref{tab:petri_net_analysis_summary_conservativeness}, the analysis of RSHPN is reduced to the analysis of PNs from the subsystem layer, PNs that form a communication model, and those that use a hybrid arrangement. However, since in the robotic system under analysis, the subsystems communicate with each other using non-blocking mode, and both the communication and transition functions are arranged in a sequential arrangement, the analysis of \pnHierarchicalXXX{}{}{} amounts to the analysis of PNs from the subsystem layer. Here only the analysis of two selected networks is presented, i.e. \pnHierarchicalXXX{\rm super, cs}{}{c} (Fig.~\ref{fig:robot_and_super_agent}) and \pnHierarchicalXXX{\rm auto, cs}{}{c} (Fig.~\ref{fig:autonomoussubsystemanalysis}), both extended accordingly. The remaining PNs are either trivial or relatively easy to analyse.

For the purpose of the analysis, the network \pnHierarchicalXXX{\rm auto, cs}{}{c} was extended to the form shown in Fig.~\ref{fig:autonomoussubsystemanalysis}.
As shown in Fig.~\ref{fig:autonomoussubsystemanalysis}, each transition has exactly one input place and exactly one output place. This means that firing any transition does not change the number of tokens in the network, hence such PN is strictly conservative.
Since in the activated network \pnHierarchicalXXX{\rm auto, cs}{}{c} the network \pnHierarchicalXXX{\rm auto}{}{} cannot insert an additional token (otherwise the safety property for the network \pnHierarchicalXXX{\rm auto}{}{} will not be fulfilled; remember that the whole \pnHierarchical{} was generated automatically following the RSHPN meta-model rules), this means that there is always exactly one token in the activated network \pnHierarchicalXXX{\rm auto, cs}{}{c} (the number of tokens is invariant, i.e. equals one). Thus the network is safe. In order to prove the absence of deadlock, it is necessary to look at both the network structure and the conditions associated with the transitions. It is enough to show that for each reachable marking (there are 7 markings because only a single token circulates among 7 places) there is always at least one transition ready to be fired (the assumption to be tested will be in fact more strict, namely whether there is exactly one transition ready to be fired).
When a~single token resides in one of the places:  \pnPlaceXXX{\rm 1}{\rm in}{}, \pnPlaceX{\rm Idle}, \pnPlaceX{\rm Termiante}, and \pnPlaceXXX{\rm 7}{\rm out}{} it is clear, that always exactly one transition is fireable, namely, \pnTransitionX{1}, \pnTransitionX{2}, \pnTransitionX{12}, and \pnTransitionX{13}, respectively.
%%%
%
\pnPlaceXXX{\rm FindBall}{}{} requires verification of the conditions associated with three transitions: \pnTransitionXXX{3}{}{} ($\pnConditionXXX{3}{}{}=[\lnot\,^cf^{\sigma}_{\rm  obstDet} \land \lnot\,^cf^{\sigma}_{\rm  term_1}]$), \pnTransitionXXX{5}{}{} ($\pnConditionXXX{5}{}{}=[^cf^{\sigma}_{\rm  obstDet} \land \lnot ^cf^{\sigma}_{\rm  term_1}]$) and \pnTransitionXXX{8}{}{} ($\pnConditionXXX{8}{}{}=[^cf^{\sigma}_{\rm  term_1}]$).
It suffices to show that when the token appears in \pnPlaceX{\rm FindBall} the following conditions are satisfied:
1)~$[\pnConditionXXX{3}{}{} \lor \pnConditionXXX{5}{}{} \lor  \pnConditionXXX{8}{}{}] = \rm True$;
2)~$\pnConditionXXX{3}{}{} \land \pnConditionXXX{5}{}{} = \rm False$;
3)~$\pnConditionXXX{3}{}{} \land \pnConditionXXX{8}{}{} = \rm False$; and
4)~$\pnConditionXXX{5}{}{} \land \pnConditionXXX{8}{}{} = \rm False$.
%%%
The logical sum of the three conditions is always $\rm True$ since: $[\lnot ^cf^{\sigma}_{\rm  obstDet} \land \lnot ^cf^{\sigma}_{\rm  term_1}] \lor [^cf^{\sigma}_{\rm  obstDet} \land \lnot ^cf^{\sigma}_{\rm  term_1}] \lor [^cf^{\sigma}_{\rm  term_1}] = [(\lnot ^cf^{\sigma}_{\rm  obstDet} \lor\, ^cf^{\sigma}_{\rm  obstDet}) \land \lnot ^cf^{\sigma}_{\rm  term_1}] \lor [^cf^{\sigma}_{\rm  term_1}] = [\lnot ^cf^{\sigma}_{\rm  term_1}] \lor [^cf^{\sigma}_{\rm  term_1}] = \rm True$.
On the other hand, logical products of consecutive condition pairs always produce False:
%%%%%
1)~$\pnConditionXXX{3}{}{} \land \pnConditionXXX{5}{}{} = [\lnot ^cf^{\sigma}_{\rm  obstDet} \land \lnot ^cf^{\sigma}_{\rm  term_1}] \land [^cf^{\sigma}_{\rm  obstDet} \land \lnot ^cf^{\sigma}_{\rm  term_1}]= [\lnot ^cf^{\sigma}_{\rm  term_1} \land (\lnot ^cf^{\sigma}_{\rm  obstDet} \land ^cf^{\sigma}_{\rm  obstDet})] = \rm False$,
%%%%%
2)~$\pnConditionXXX{3}{}{} \land \pnConditionXXX{8}{}{} = [\lnot ^cf^{\sigma}_{\rm  obstDet} \land \lnot ^cf^{\sigma}_{\rm  term_1}] \land [^cf^{\sigma}_{\rm  term_1}] = \rm False$ (since $\lnot ^cf^{\sigma}_{\rm  term_1} \land\, ^cf^{\sigma}_{\rm  term_1} = \rm False$) and,
3)~$\pnConditionXXX{5}{}{} \land \pnConditionXXX{8}{}{} = [^cf^{\sigma}_{\rm  obstDet} \land \lnot ^cf^{\sigma}_{\rm  term_1}] \land [^cf^{\sigma}_{\rm  term_1}] = \rm False$ (since $\lnot ^cf^{\sigma}_{\rm  term_1} \land\, ^cf^{\sigma}_{\rm  term_1} = \rm False$).
%%%%%%%%%%%%%%%%%%%%%%%
%%%%%%%%%%%%%%%%%%%%%%%
%%%%%%%%%%%%%%%%%%%%%%%
An analogous analysis should be performed for the other two places, namely \pnPlaceX{\rm AvoidObstacle} and \pnPlaceX{\rm CollectBall}. In the case where the token appears in \pnPlaceX{\rm AvoidObstacle} only one transition is ready to fire because the conditions associated with the transitions \pnTransitionX{6} and \pnTransitionX{9}, i.e. $\pnConditionX{6}=[\lnot ^cf^{\sigma}_{\rm  term_2}]$, and $\pnConditionX{9}=[^cf^{\sigma}_{\rm  term_2}]$, respectively, are mutually exclusive.
On the other hand, three transitions are enabled when a~token appears in \pnPlaceX{\rm CollectBall}, namely: $\pnTransitionX{4}$ ($\pnConditionX{4}=[\lnot ^cf^{\sigma}_{\rm  obstDet} \land \lnot ^cf^{\sigma}_{\rm  term_3}]$), $\pnTransitionX{7}$ ($\pnConditionX{7}=[^cf^{\sigma}_{\rm  obstDet} \land \lnot ^cf^{\sigma}_{\rm  term_3}]$), and $\pnTransitionX{10}$ ($\pnConditionX{10}=[^cf^{\sigma}_{\rm  term_3}]$).
Proving that: 1) $\pnConditionX{4}\lor\pnConditionX{7}\lor\pnConditionX{10}$, 2) $\pnConditionX{4}\land\pnConditionX{7}$, 3) $\pnConditionX{4}\land\pnConditionX{10}$ and 4) $\pnConditionX{7}\land\pnConditionX{10}$, is straightforward enough to be omitted. This leads to the final conclusion that for each reachable marking in the extended \pnHierarchicalXXX{\rm auto, cs}{}{c} always at least a~single transition is fireable, i.e.\ the \pnHierarchicalXXX{\rm auto, cs}{}{c} is deadlock-free.

The analysis of the extended \pnHierarchicalXXX{\rm super,cs}{}{c} (Fig.~\ref{fig:robot_and_super_agent}) is carried out in a~similar way. In the extended \pnHierarchicalXXX{\rm super, cs}{}{c}
each transition has exactly one input place and exactly one output place. Consequently, the network is strictly conservative.
Considering that and the fulfilment of the safety property by \pnHierarchicalXXX{\rm super}{}{}, in the extended \pnHierarchicalXXX{\rm super, cs}{}{c}
there can be exactly one token. This leads to the conclusion that \pnHierarchicalXXX{\rm super,cs}{}{c} is safe.
As there can be only one token circulating in \pnHierarchicalXXX{\rm super,cs}{}{c}, the number of reachable markings equals to the number of places, i.e.\ 8.
In the case where a~single token resides in one of the six places, i.e. \pnPlaceXX{1}{\rm in}, \pnPlaceX{\rm TeleActivated}, \pnPlaceX{\rm AutoActivated}, \pnPlaceX{\rm StopRobot}, \pnPlaceX{\rm Termination}, \pnPlaceXX{8}{\rm out}, then exactly one transition
is fireable (because each place has exactly one output transition, and each of output transitions has its condition set to the default value, i.e. $\rm True$):
\pnTransitionX{1}, \pnTransitionX{5}, \pnTransitionX{6}, \pnTransitionX{7}, \pnTransitionX{8} and \pnTransitionX{10}, respectively. In the case of place \pnPlaceX{\rm Idle}, the conditions associated with its output transitions: \pnTransitionX{2} and \pnTransitionX{9}, are mutually exclusive (since $\pnConditionX{2}=[^cf^{\sigma}_{\rm taskActiv}]$ and $\pnConditionX{9}=[\lnot ^cf^{\sigma}_{\rm taskActiv}]$). Similarly, in the case of place \pnPlaceX{\rm Activation}, the conditions associated with its output transitions: \pnTransitionX{3} and \pnTransitionX{4}, are mutually exclusive (since $\pnConditionX{3}=[\lnot ^cf^{\sigma}_{\rm autoActivated}]$ and $\pnConditionX{4}= [^cf^{\sigma}_{\rm autoActivated}]$).
This leads to the final conclusion that for each reachable marking of the extended \pnHierarchicalXXX{\rm super, cs}{}{c} there is always exactly one fireable transition. This proves that the network \pnHierarchicalXXX{\rm super, cs}{}{c} is deadlock-free.
%%%%

Taking into account the above considerations, it can be concluded that the network \pnHierarchicalX{} is safe, deadlock-free and it satisfies the conservativeness assumptions presented in Table~\ref{tab:petri_net_analysis_summary_conservativeness}.

%%%%%%%%%%%%%%%%%%%%%%%%%%

% !TeX root = ../figat_analiza.tex

\section{Discussion}
\label{sec:discussion}

\subsection{Selected methods for property verification}
\label{subsec:selected-methods-for-property-verification}
%{\bf Selected methods for property verification:}
In this paper, we have selected three methods for analysing PNs, focusing on their simplicity and applicability to specific layers of the RSHPN model. These methods include graphical network state analysis (simulating the network), reachability graph analysis, and analysis based on place and transition invariants. They were chosen because they sufficiently verify key system properties such as safety, conservatism, and deadlock freedom. However, other methods, such as critical path analysis or temporal analysis, could also be used to explore additional properties. Future enhancements could include incorporating temporal analysis for tasks or communication models, and combining model checking with runtime verification by leveraging the meta-model checking approach.

Our primary goal was to develop a method to decompose the analysis into manageable parts. For each RSHPN layer, we specify the most appropriate analysis method and identify which layers do not require repeated analysis. For example, general patterns, such as the behaviour or communication model layers, only need to be analysed once. Networks generated by populating the RSHPN meta-model with the relevant parameters will retain the properties of these layers, avoiding redundant verification.
The proposed methods are currently adequate for verifying system determinism and deadlock freedom. The decomposition method introduced here can also serve as a basis for analysing other properties in future studies.

\begin{figure}
	\begin{subfigure}{0.4\linewidth}
		\caption{}
		\label{fig:block-block-example}
		\includegraphics[width=0.7\linewidth]{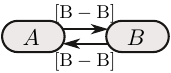}
		\includegraphics[width=1.0\linewidth]{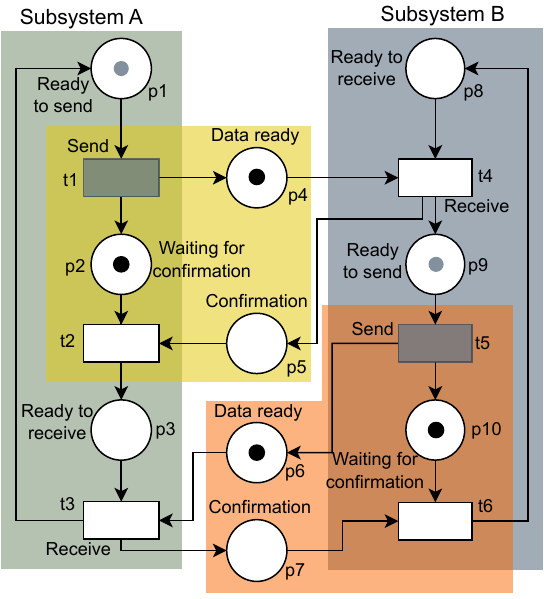}

	\end{subfigure}
	\begin{subfigure}{0.57\linewidth}
		\caption{}
		\label{fig:local-deadlock-example}
		\centering
		\includegraphics[width=1.0\linewidth]{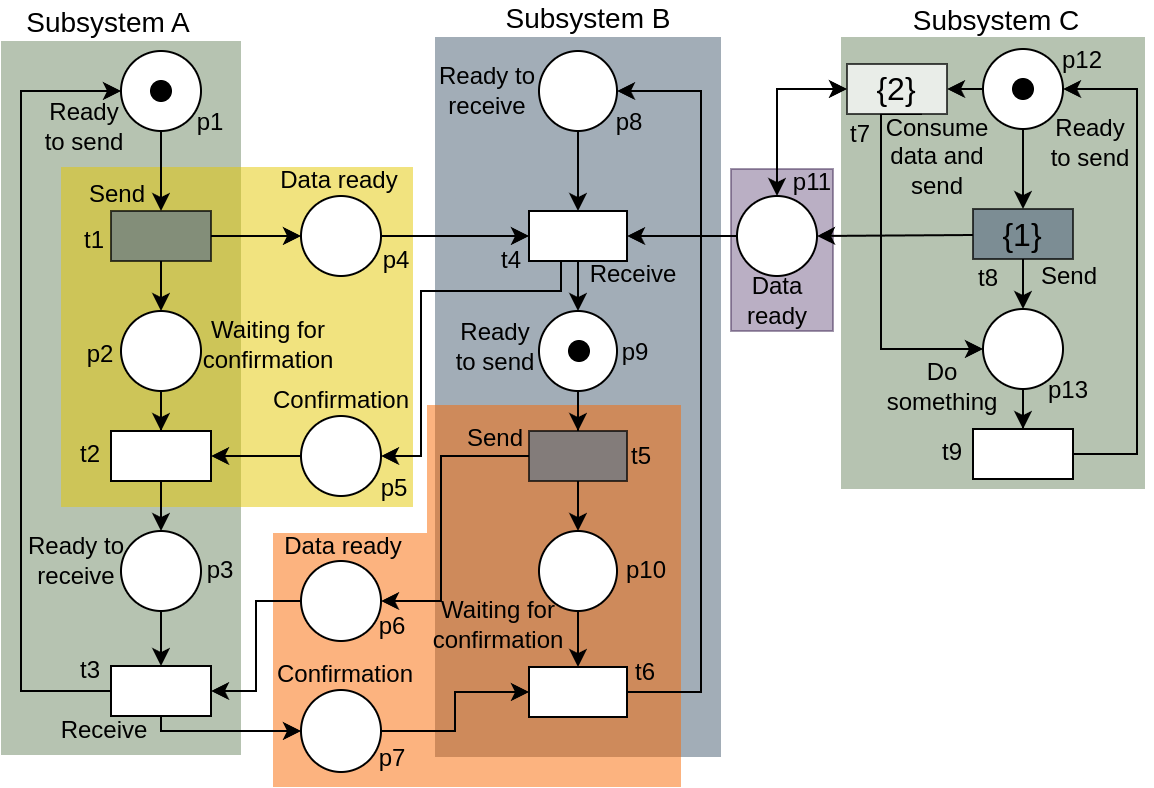}\\
		\includegraphics[width=1.0\linewidth]{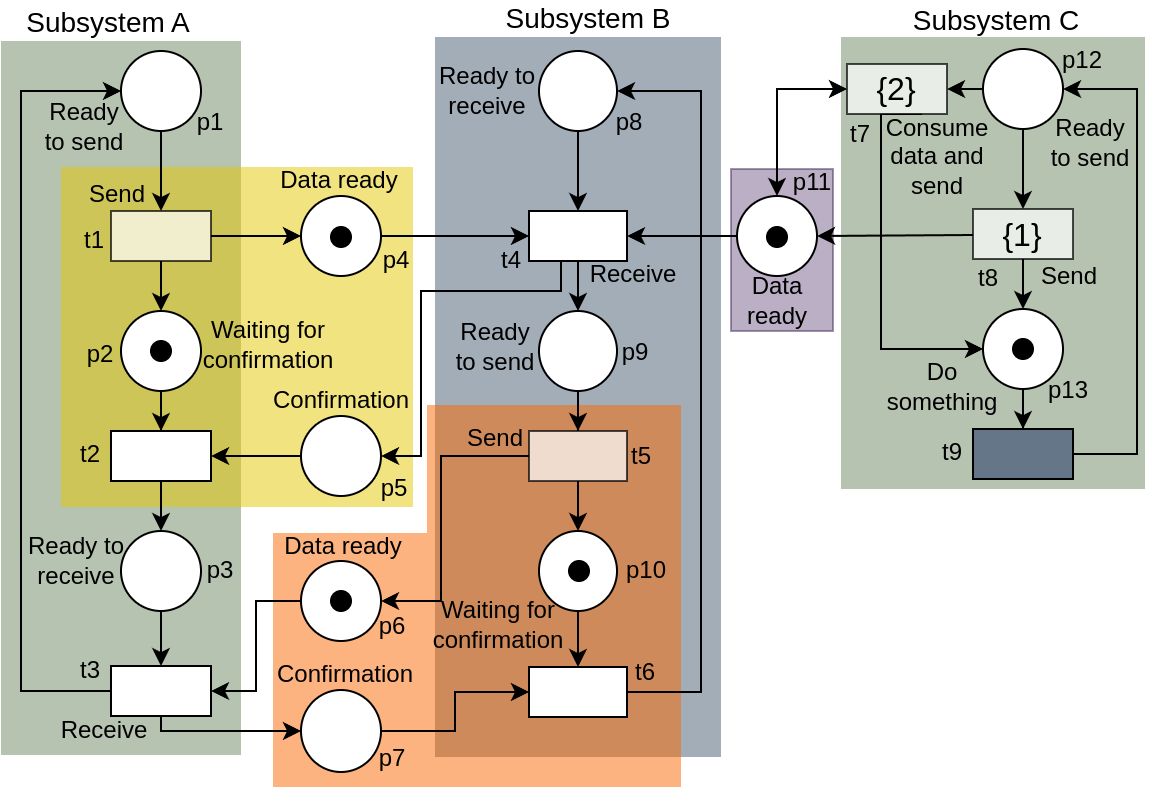}	
	\end{subfigure}

	\caption{{\bf (a)}: Subsystems A and B send and receive data in blocking mode. If A starts with a token in p1 and B in p9 then the deadlock occurs. However, if they are appropriately initialised, i.e.\ A starts with p1 and B with p10 then there is no deadlock; {\bf (b):} Example of a local deadlock, without global deadlock. Subsystem A and B have no transition to fire, while subsystem C has. Right top: initial marking, Right bottom: occurrence of local deadlock (after firing: t1, t5, t8, t9, t7).
	The priority of t7 (\{2\}) is higher than that of t8 (\{1\}).}
	\label{fig:block_deadlock_example_and_local_deadlock_example}
\end{figure}

\subsection{Different causes of deadlocks}
\label{subsec:different-causes-of-deadlocks}
%{\bf Different causes of deadlocks:}
Sec.~\ref{sec:rshpn_model_analysis} discusses the analysis of properties of the RSHPN meta-model, concentrating on potential causes of deadlocks. While there may be many causes, we underscore two key ones, that should be considered when designing the RSHPN: (1) deadlocks occurring due to subsystem activities, where a lower-level PN blocks the token and prevents it from reappearing in the higher-level network, and (2) inappropriate system initialization or incorrect selection of communication modes, e.g.\ inappropriate use of blocking modes (as in Fig.~\ref{fig:block-block-example}).
	
The absence of a deadlock in an RSHPN (no global deadlock) does not mean that there is no local deadlock, i.e.\ that a subsystem or agent, or communication between several subsystems, is blocked by another agent/subsystem running in parallel. Consider the example shown in Fig.~\ref{fig:local-deadlock-example}, where there is a local deadlock (between subsystems A and B - bottom of Fig.~\ref{fig:local-deadlock-example}), while there is no global deadlock, i.e.\ there is always at least one transition in the network (in subsystem C) ready to fire, but the network models 3 subsystems running in parallel. It is therefore clear that the analysis methods we propose can be helpful in ensuring the correct operation of a robotic system.

\subsection{Communication in multi-robotic systems}
\label{subsec:communication-in-multi-robotic-systems}
In multi-robot systems, effective communication is of paramount importance to ensure proper coordination between subsystems. While this study focused on the analysis of blocking, non-blocking, and blocking with timeout communication modes, it is important to acknowledge that other aspects are equally significant, as evidenced by~\cite{Zielinski-2017-JINT:new}. This is consistent with the findings of~\cite{MultiModalCommunicationBinaryDecomposition:2022} concluding that there is a need for further investigation into the role of message content in multi-robotic communication. It is important to note that the content exchanged between agents depends on the specific task, and therefore cannot be generalised within the RSHPN meta-model, which is a generic template, i.e.\ the task is abstracted away.

Our approach, based on the RSHPN meta-model, allows not only 1-to-1 communication, but also 1-to-many communication, i.e. broadcasting, where receivers may be known or unknown to the sender. In cases where the receivers are unknown, communication takes place via stigmergy, where subsystems communicate indirectly via the environment. This is particularly relevant in swarm systems, where robots interact locally without direct coordination. In such cases, separate RSHPNs are constructed for each robot.

\subsection{Modelling and analysis of multi-robot systems}
\label{subsec:modeling-and-analysis-of-multi-robotic-systems}

\mfb{A multi-robot system consists of many robots working together to achieve a common objective. These systems can be categorized based on their coordination mechanisms: coordinated ones having a coordinator or swarm systems that rely on decentralized interactions.}

\mfb{{\bf Coordinated Multi-Robot Systems:}
In coordinated multi-robot systems, synchronisation of activities is typically achieved through predefined communication channels and explicit task allocation done by the coordinator. Such systems can be modelled either through separate RSHPNs for each robot or a single RSHPN with fused places that synchronize agents and subsystems at their action layers.
The latter situation is similar to the one present in the ping-pong ball collecting robot system~\cite{Figat:2022:RAS}, where a variable structure was used, allowing for the dynamic reconfiguration of subsystems by the coordinator.}

\mfb{{\bf Swarm Robot Systems:}
Swarm robot systems do not have a coordinator and they aspire to exhibit an emergent behaviour which results from their interaction.  Usually those robots interact with minimal direct communication and base their actions on the observation of the environment. These systems exhibit self-organization and robustness, and are often inspired by biological swarms such as bees or ants \cite{Gordon:2010}. The coordination in swarm robotics is typically achieved through indirect communication, such as stigmergy, where robots modify their environment to influence the behaviour of others.}

\mfb{In a swarm robot system, each robot can be modelled as an independent RSHPN, but their interactions emerge from environmental changes rather than direct exchanges of information. Therefore, it is unnecessary to analyse two PNs simultaneously if robots do not directly communicate. However, if members of the swarm communicate, their networks merge at the lowest layer, i.e., the sixth one, responsible for communication, and the first one, which initializes all agents. In nature, social insects such as ants~\cite{Gordon:2010} primarily interact using stigmergy. However, in rare cases, ants communicate directly via antennae interactions. So in this case individual RSHPNs for each robot would have to be analysed.}

\mfb{Nevertheless, integrating multiple networks could be considered if the environment in which the robots operate was modelled as a PN. If one robot’s actions affected the state of the environment, which was itself represented as a PN, and another robot could sense and respond to this change accordingly, then merging these networks would be meaningful. This suggests an interesting future research direction focused on the emergence of swarm behaviours in robotic systems. The ability to integrate the environment PN model into the RSHPN would enable the utilisation of the proposed analysis method. Thus both coordinated multi-robot systems and robotic swarms could be analysed. However, environment modelling PN wold have to possesses similar structure to the ones composing the  RSHPN. Ideally, such a network should have a single input place and a single output place, ensuring simplified property analysis. This approach guarantees that environmental changes introduced by an individual robot can be detected by others in its vicinity. In the case of a coordinated system, this would lead to the formation of a single large network encompassing all agents at the multi-agent robotic system layer. An intriguing question is whether such a network could provide insights into system-level emergent behaviours. Investigating this possibility is certainly worthwhile, but it lies beyond the scope of this paper.}

% !TeX root = ../figat_analiza.tex

%\clearpage

\section{Conclusions}
\label{sec:conclusions}

The RSHPN meta-model simplifies the verification of the correctness of a PN modelling the activities of a specific \roboticSystem. The meta-model serves as a template into which designed system specific parameters are inserted. The meta-model is thus transformed into a \roboticSystem\ model. This can be done using the RSSL language \cite{Figat:2022:RAL}. The RSHPN decomposition into 6 layers and the separation of panels defining PNs with one input and one output place avoid PN state space explosion and scalability issues. This enables network analysis to use reduction methods that collapse higher order networks into single transitions and lower order networks into single places, while preserving their properties.
Key parts of the meta-model are general (e.g., behaviour layer PNs use the same pattern; multi-agent robotic systems and agent layers form trivial PNs) and are analysed once for all systems. Thus, the analysis of a specific \roboticSystem\ model is limited to a few PNs, such as subsystem layer PNs or communication and computation composition PNs. This allows designers to focus on analysing of the critical aspects of the system, such as the task definition and communication model. Modification of systems designed by using RSHPN is much easier (as indicated in~\cite{Figat:2022:RAS,Figat:2022:RAL}), because of the modular structure only some subsystems need to be replaced, and therefore only new PNs need to be created and analysed.

Our motivation is to present a general meta-model that can be applied to the design of practically any robotic system, whether it is a single or multi-robot systems, with fixed or variable structures. In particular,  the paper emphasises the formal analysis that can reveal system properties already in the specification phase. Verifying these properties prior to implementation is crucial, as it prevents costly errors that would otherwise be found during testing. Whilst the majority of PNs in the RSHPN meta-model satisfy the required properties, task-specific PNs should be developed according to the guidelines provided, offering significant value to system designers.

The PN analysis method in the future can be further extended to investigate additional \roboticSystem\ properties: (1) model checking (e.g.: schedulability, reachability~\cite{Su:2023}, admissibility~\cite{Lacerda:2019}) and (2) runtime verification (e.g.: task period and WCET overshoot~\cite{DalZilio:2023}). Combining model checking with runtime verification, either by restricting the analysed state space~\cite{Pelletier:2023} or by introducing predictive runtime verification~\cite{Pinisetty:2017} using the system model as \emph{a~priori\/} knowledge, would be beneficial. The RSHPN could be extended with constraints using Linear Temporal Logic (LTL) to introduce a supervisor to monitor the system, as in~\cite{Lacerda:2019,Pelletier:2023}, and to propose tools to optimise system resources~\cite{Lahijanian:2018}, mission, or tasks.
 % OK

% !TeX root = ../figat_analiza.tex

   % OK

\vfill

\end{document}